\documentclass[hyperfootnotes=false]{article}
\usepackage{amsmath,graphicx}
\usepackage{subfig}
\usepackage{tabu}
\usepackage{rotating}

\usepackage[table]{xcolor}
\usepackage{collcell}
\usepackage{hhline}
\usepackage{pgf}
\usepackage{url}

\newcommand\blfootnote[1]{%
	\begingroup
	\renewcommand\thefootnote{}\footnote{#1}%
	\addtocounter{footnote}{-1}%
	\endgroup
}

\begin{document}
	
\begin{Huge}
IEEE Copyright Notice
\end{Huge}

\vspace{\baselineskip}

Copyright (c) 2016 IEEE

Personal use of this material is permitted.
Permission from IEEE must be obtained for all other	uses, in any current or future media, including reprinting/republishing this material for advertising or promotional purposes, creating new collective works, for resale or redistribution to servers or lists, or reuse of any copyrighted component of this work in other works.

\vspace{\baselineskip}

C. Gacav, B. Benligiray, and C. Topal, “Greedy search for descriptive spatial face features,” in Proc. IEEE Int. Conf. Acoust., Speech and Signal Process. (ICASSP), 2017, pp. 1497--1501.

\vspace{\baselineskip}

\url{http://dx.doi.org/10.1109/ICASSP.2017.7952406}

\newpage
	
\title{Greedy Search for Descriptive Spatial Face Features}
	
\author{Caner Gacav \\ canergacav@gmail.com \and Burak Benligiray \\ burakbenligiray@anadolu.edu.tr \and Cihan Topal \\ cihant@anadolu.edu.tr}

\date{}
	
\maketitle

\begin{abstract}
Facial expression recognition methods use a combination of geometric and appearance-based features.
Spatial features are derived from displacements of facial landmarks, and carry geometric information.
These features are either selected based on prior knowledge, or dimension-reduced from a large pool.
In this study, we produce a large number of potential spatial features using two combinations of facial landmarks.
Among these, we search for a descriptive subset of features using sequential forward selection.
The chosen feature subset is used to classify facial expressions in the extended Cohn-Kanade dataset~(CK+), and delivered 88.7\% recognition accuracy without using any appearance-based features.
\blfootnote{Source code: \url{https://github.com/bbenligiray/greedy-face-features}}
\end{abstract}

\section{Introduction}
\label{sec:intro}

Facial expressions are important cues that support verbal communication.
Analyzing individuals' psychological states and emotions by their facial expressions has become widespread in human behavior analysis and human--computer interaction studies~\cite{Ekman:1997,Bartlett:2003}.
Automated computer vision methods that gather facial expression data allow these studies to be conducted more effortlessly~\cite{Zeng:2009,Rehg:2013}.
As the technology advances, vision systems will be able to sense subtle emotions and sentiments that humans cannot.

Geometric and appearance-based features are commonly used in facial expression recognition.
In this study, we focus on spatial features, which are a type of geometric feature.
Spatial features are calculated using the displacements of a combination of facial landmarks.
Due to the high number of combinations, there are many potential spatial features, of which some are more descriptive.
To provide the classifier with a descriptive subset of features with little redundancy, selection can be made based on prior knowledge~\cite{Tian:2001,Suk:2014}.
For example, Facial Action Coding System (FACS) defines a set of Action Units that produce expressions~\cite{Ekman:1997}.
Another approach is to explicitly apply dimension reduction~\cite{Huang:2014,Ekenel:2004} or let the classifier handle the selection~\cite{Chen:2015}.
As an alternative to these methods, we use a feature selection algorithm to form a descriptive feature subset.

Two combinations of 68 facial landmarks result in 2278 landmark pairs.
We handle horizontal and vertical distance variations between these landmark pairs as separate features, thus work with 4556 potential features.
Forward sequential feature selection reduces the number of features to 7.
The resulting subset of features is used for classification in the extended Cohn-Kanade dataset~(CK+)~\cite{Lucey:2010}.
By using the selected spatial features, 88.7\% recognition accuracy is obtained.
This result surpasses other methods using only geometric features, and can be improved by utilizing appearance-based features.

\section{Related Work}

Feature extraction and classification for facial expression recognition is a well-established problem in computer vision~\cite{Pantic:2000,Fasel:2003}.
Huang et al.\ use local binary patterns as appearance based features~\cite{Huang:2014}.
They build a canonical subspace of the subsequent frames, and model the lower-dimensional feature space using discriminative canonical correlation analysis.

Lucey et al.\ detect facial landmarks using active appearance models~\cite{Lucey:2010}.
These landmarks are used to calculate similarity-normalized shape~(SPTS) and canonical appearance~(CAPP) features.
Suk and Prabhakaran locate facial landmarks using an active shape model, and use displacements between landmarks located from neutral and expressive faces as features~\cite{Suk:2014}.

Chen et al.\ use appearance-based and geometric features~\cite{Chen:2015}.
Appearance-based features are represented by histogram of gradients (HOG) from three orthogonal planes. 
Geometric features are divided into two categories, namely rigid and non-rigid changes.
Multiple kernel learning is used to find an optimal combination of these features.
Turan and Lam extract features from the eye and mouth regions using local phase quantization and pyramid of HOG descriptors~\cite{Turan:2014}.
Features are fused using canonical correlation analysis and classified with SVM.

In our previous work, we used the variations in Euclidean distances between landmark pairs as spatial features~\cite{Gacav:2016}, which gives slightly worse results than handling horizontal and vertical distances independently~\cite{Lucey:2010}.
Leave-one-subject-out was used instead of 10-fold cross-validation, producing optimistic results due to CK+ dataset providing small number of examples for some classes.

Deep learning methods have grown to be an important part of literature for all computer vision problems.
However, the modest sizes of current datasets may be limiting their prevalence in facial expression recognition~\cite{Lucey:2010,Lyons:1998}.
Deep learners optimize feature design, feature selection and classification steps jointly.
Liu et al.\ design a deep belief net that trains for these steps iteratively~\cite{Liu:2014}.

\section{Proposed Method}

\begin{figure}
	\captionsetup[subfigure]{labelformat=empty}
	\centering	
	\subfloat{{\includegraphics[width=0.2\textwidth]{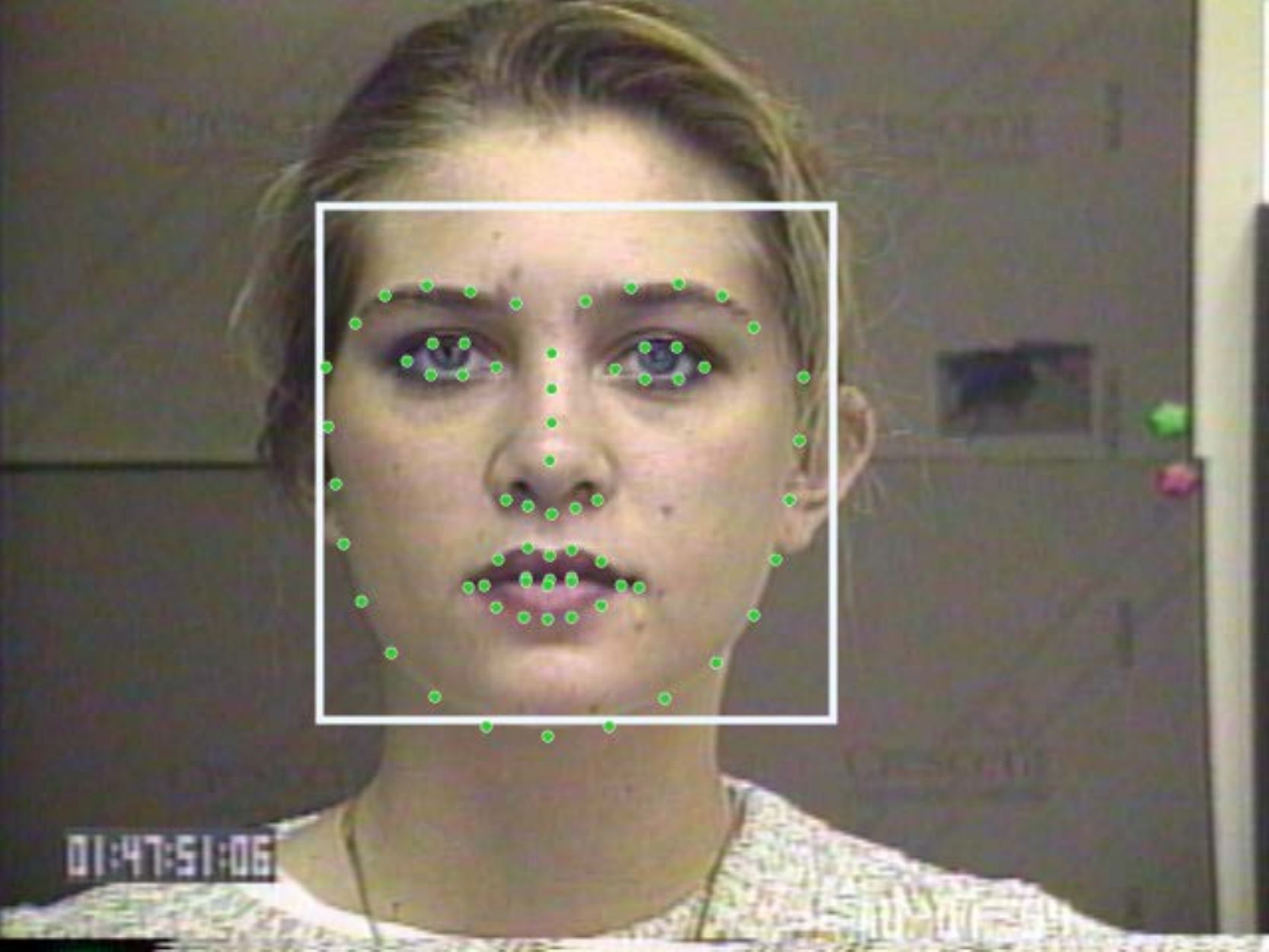}}}
	\subfloat{{\includegraphics[width=0.2\textwidth]{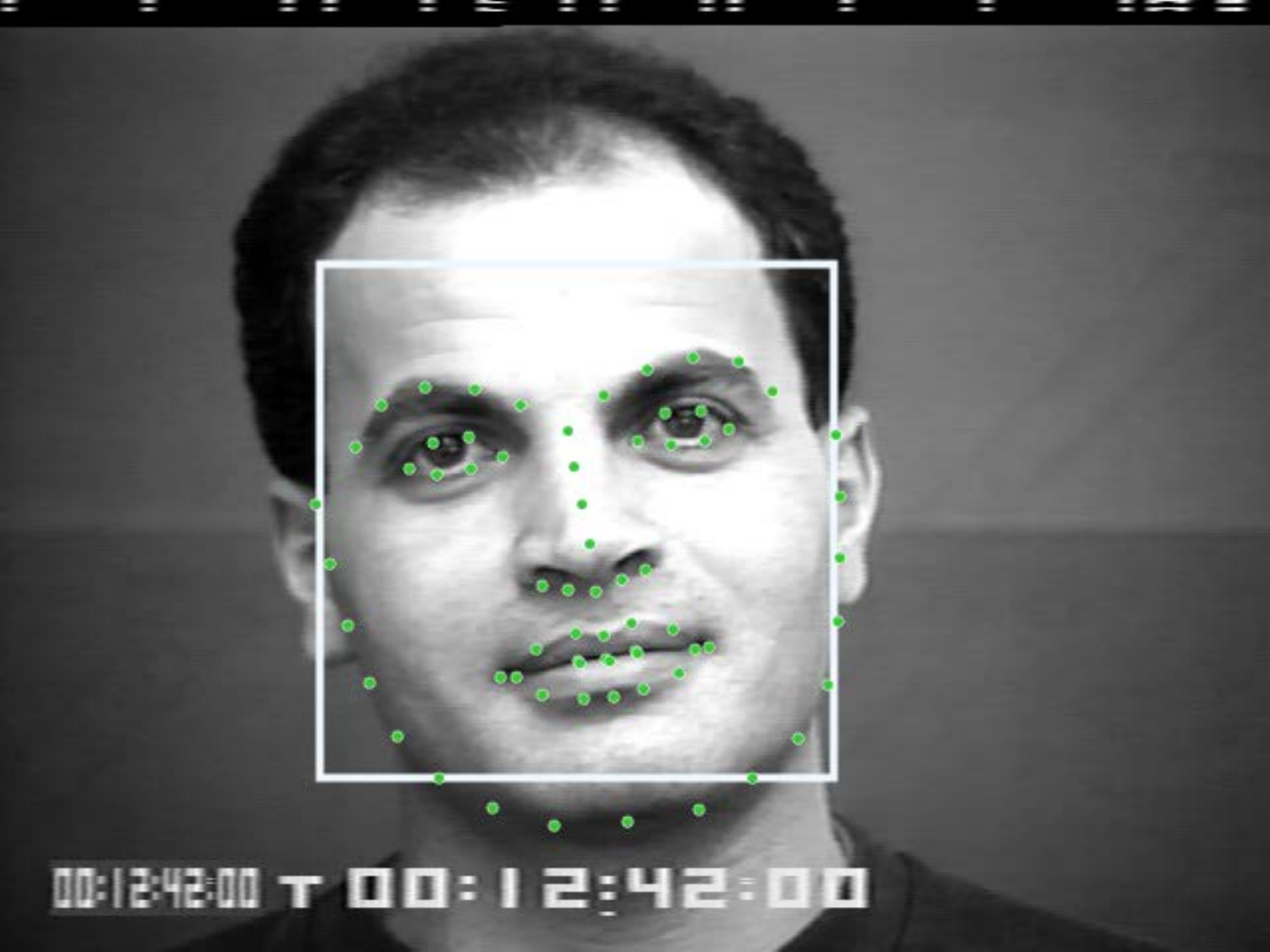}}}
	\subfloat{{\includegraphics[width=0.2\textwidth]{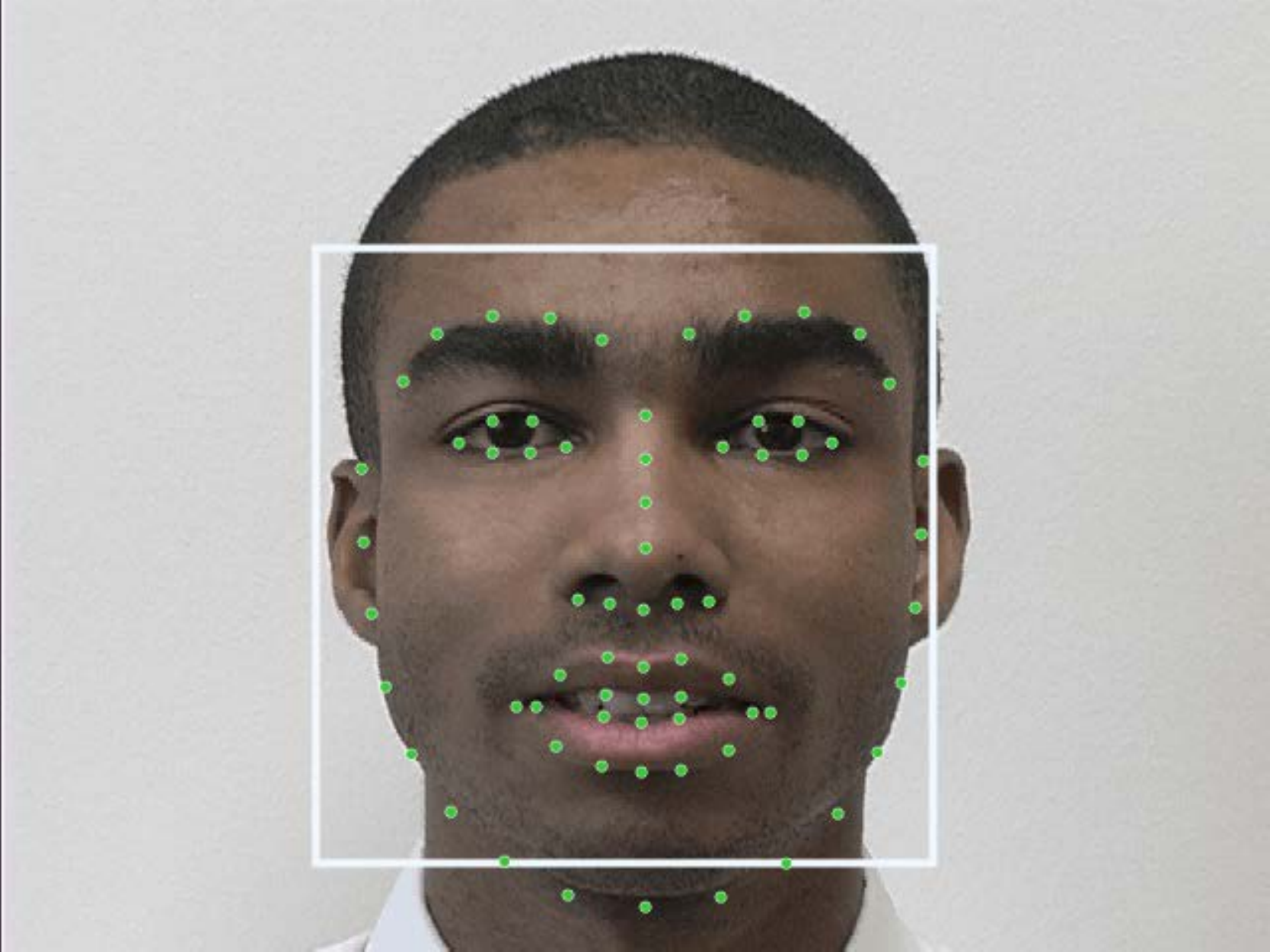}}}
	\subfloat{{\includegraphics[width=0.2\textwidth]{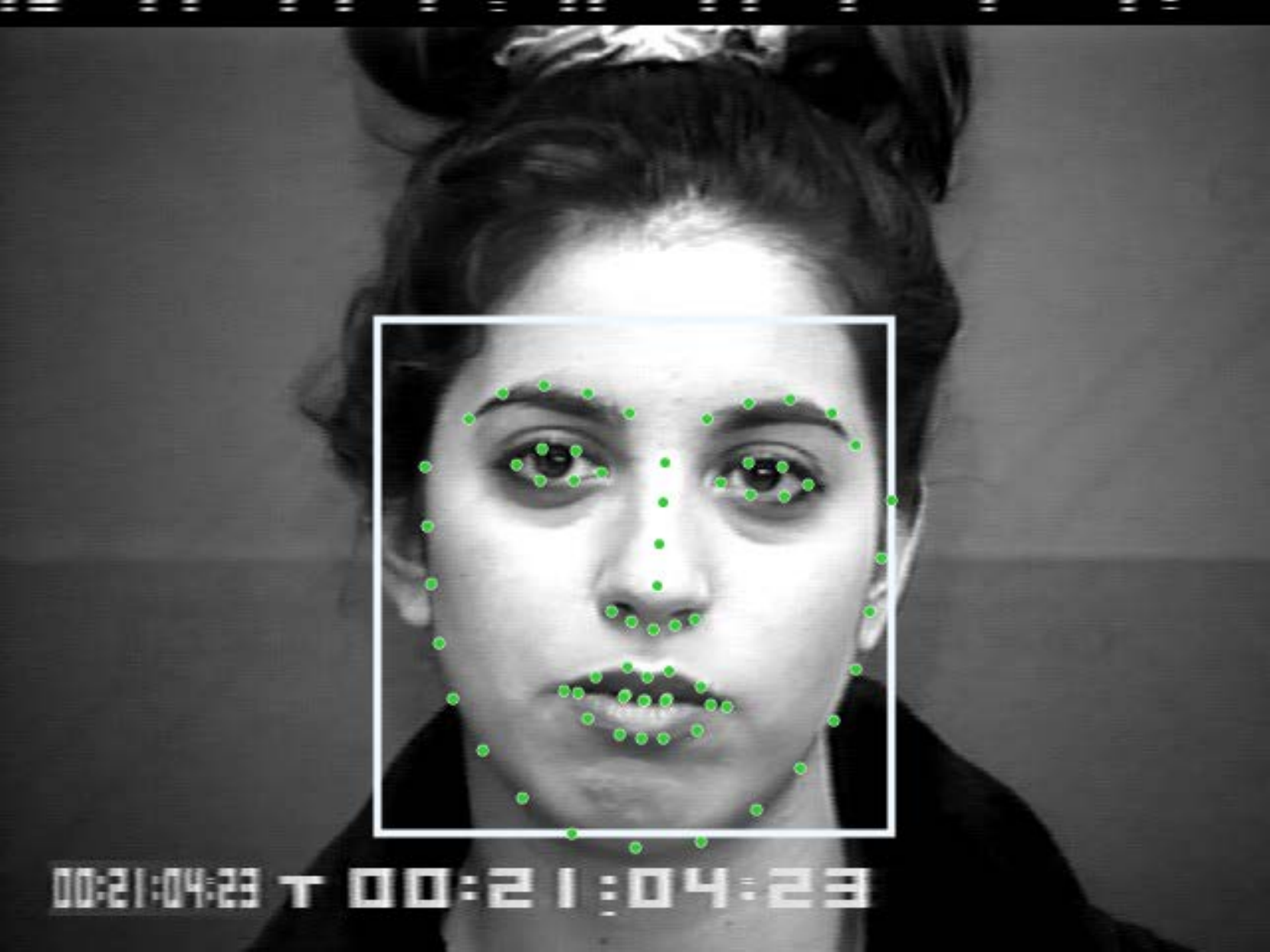}}}
	\subfloat{{\includegraphics[width=0.2\textwidth]{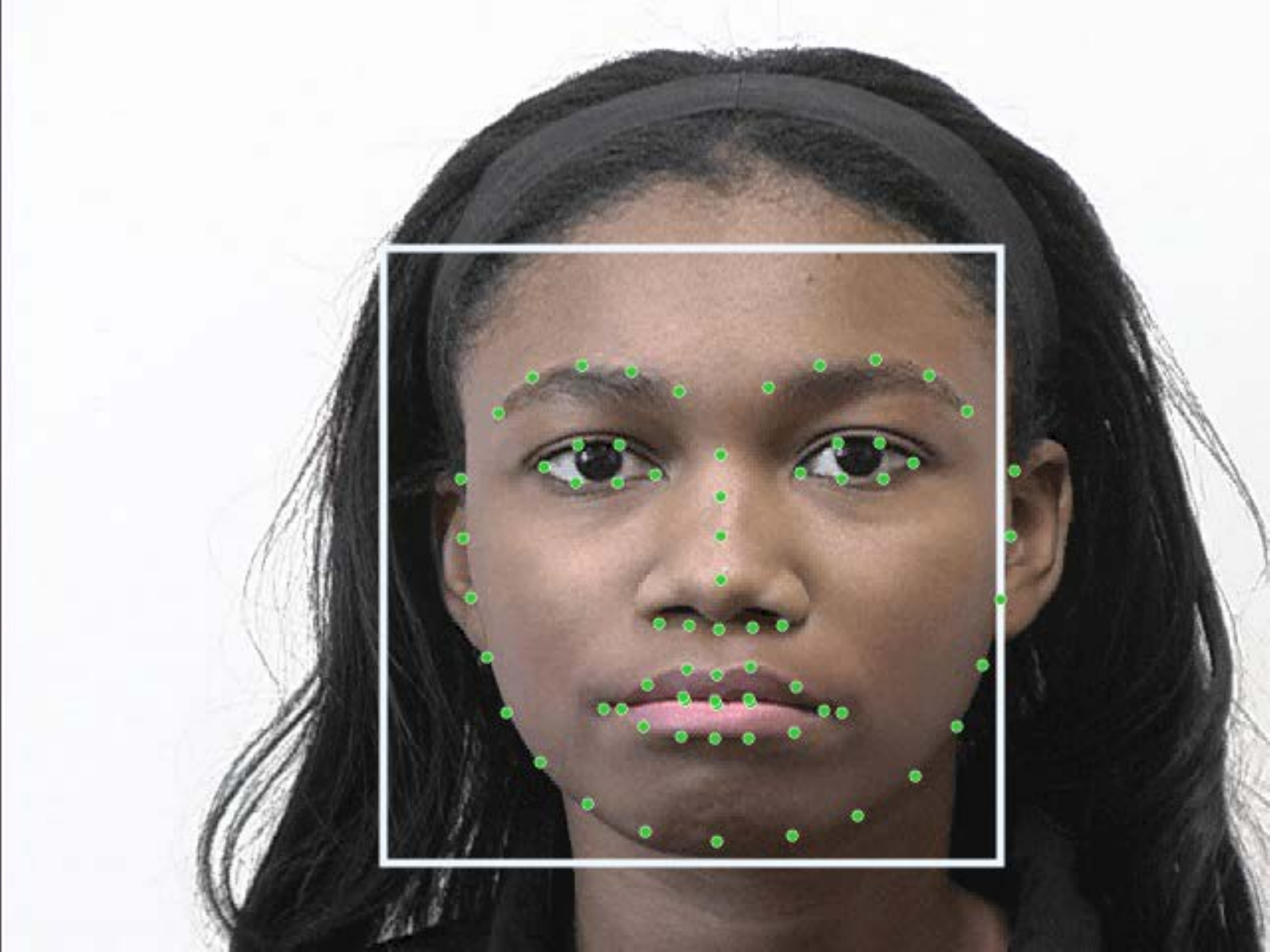}}}
	\vspace{-3mm}
	\subfloat[Contempt]{{\includegraphics[width=0.2\textwidth]{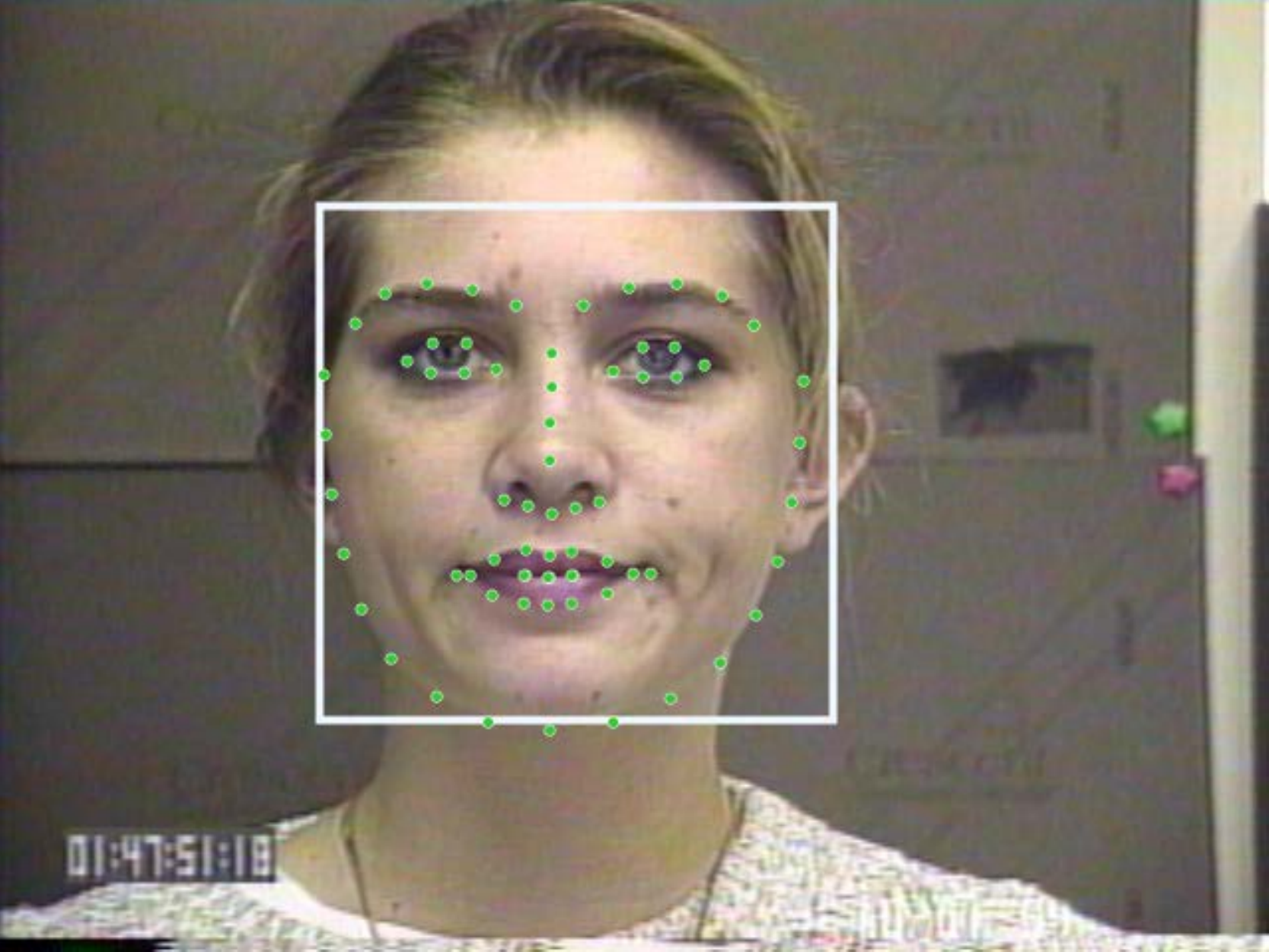}}}
	\subfloat[Disgust]{{\includegraphics[width=0.2\textwidth]{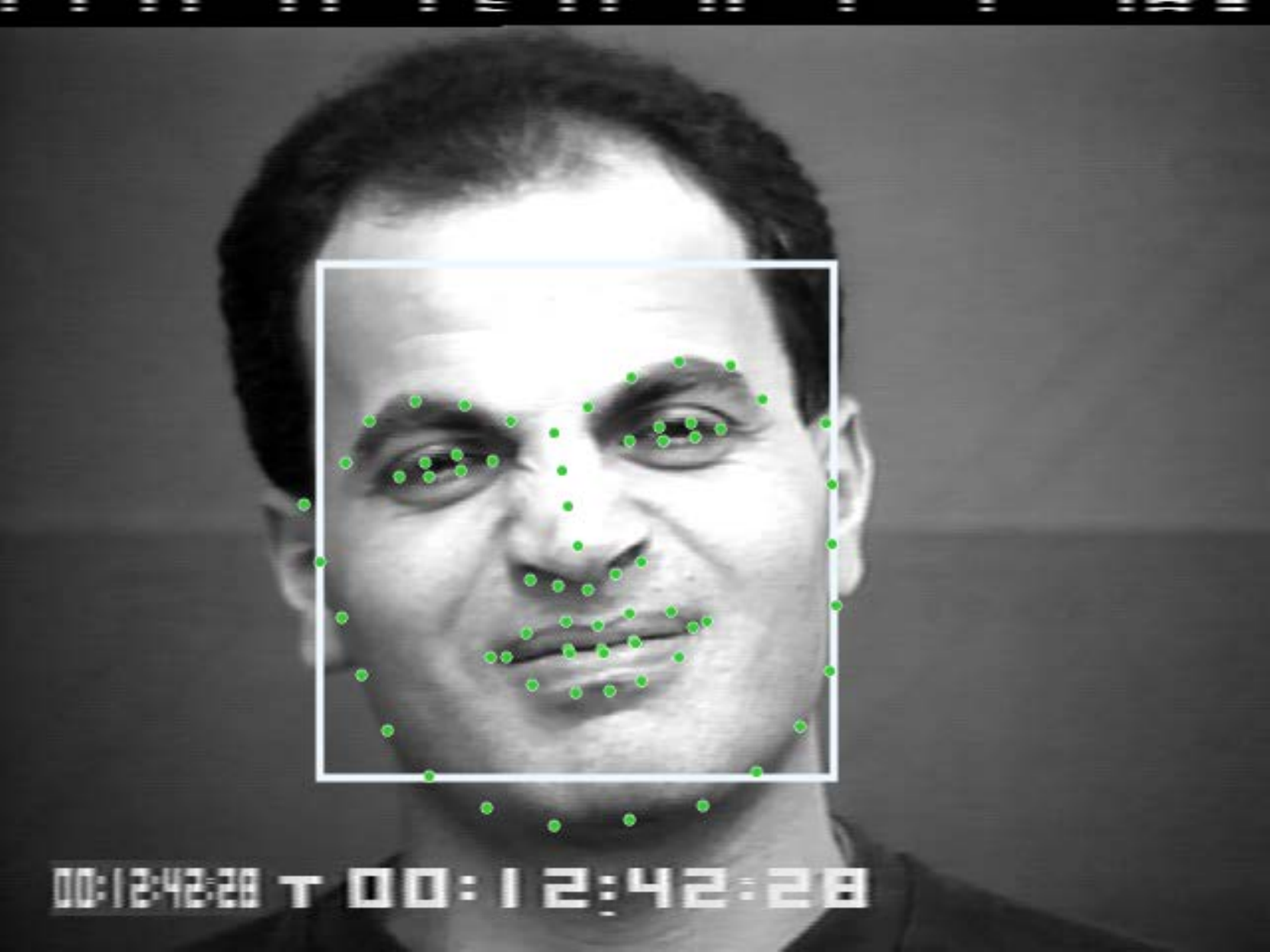}}}
	\subfloat[Fear]{{\includegraphics[width=0.2\textwidth]{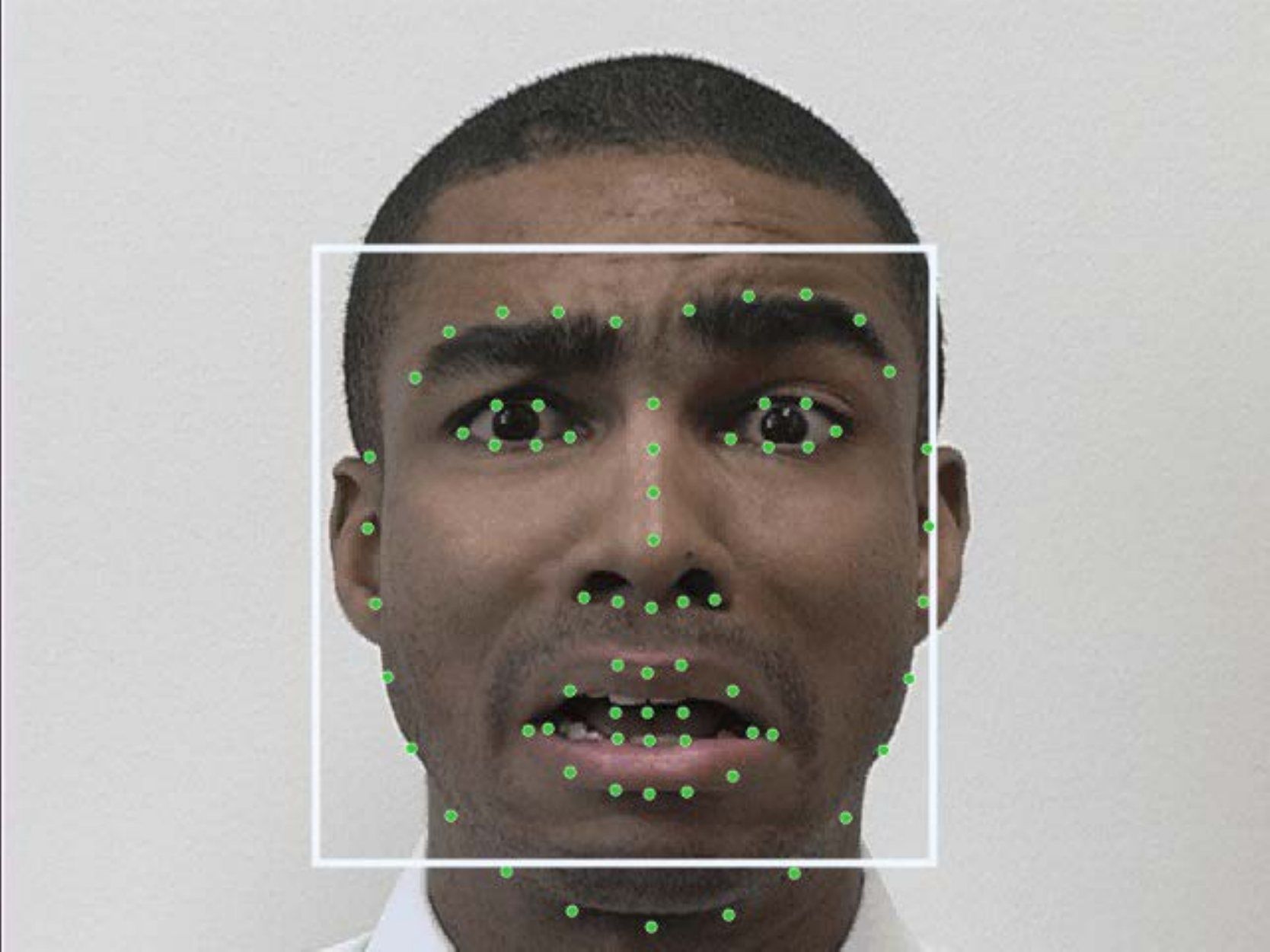}}}
	\subfloat[Happiness]{{\includegraphics[width=0.2\textwidth]{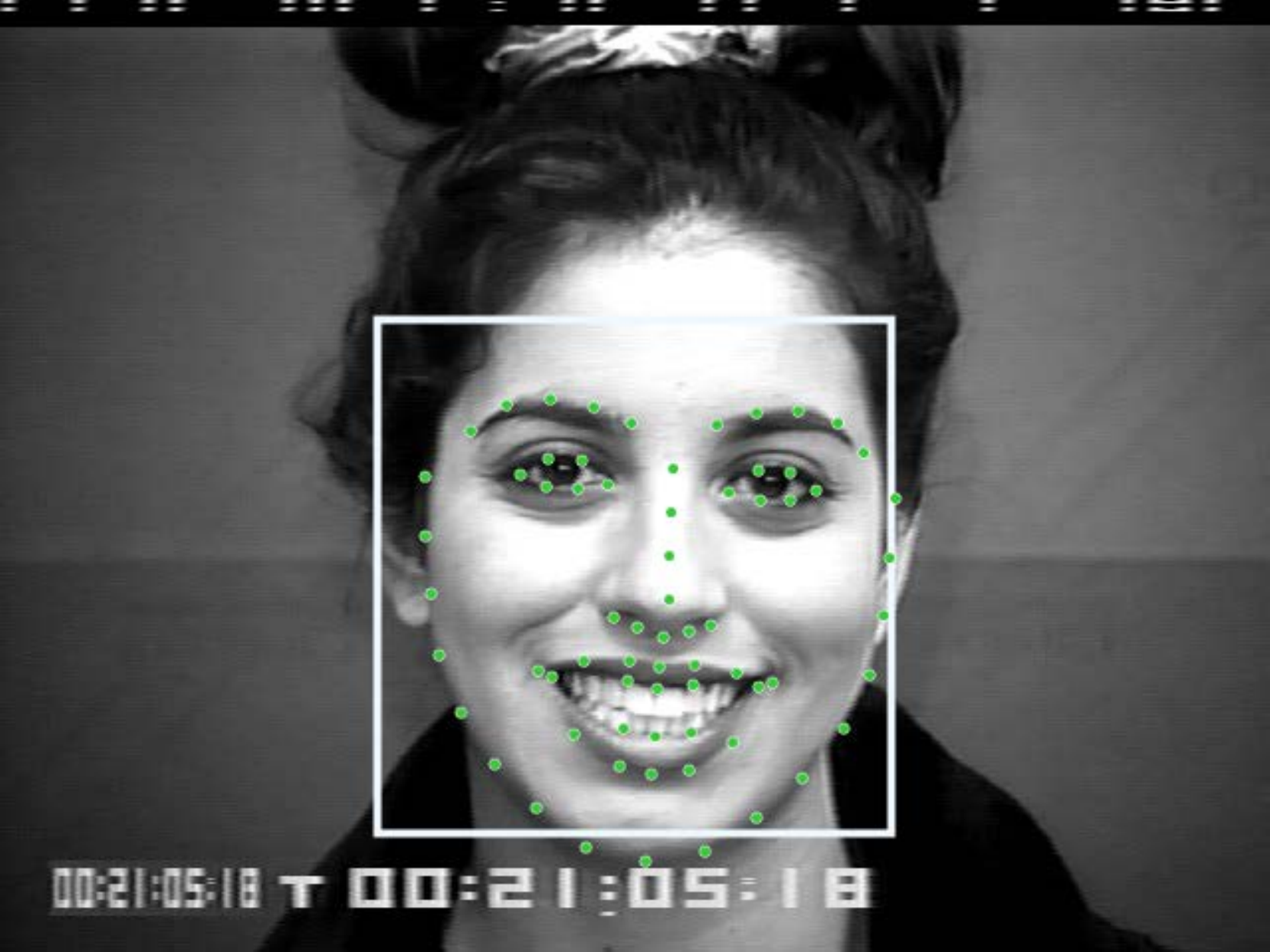}}}		
	\subfloat[Sadness]{{\includegraphics[width=0.2\textwidth]{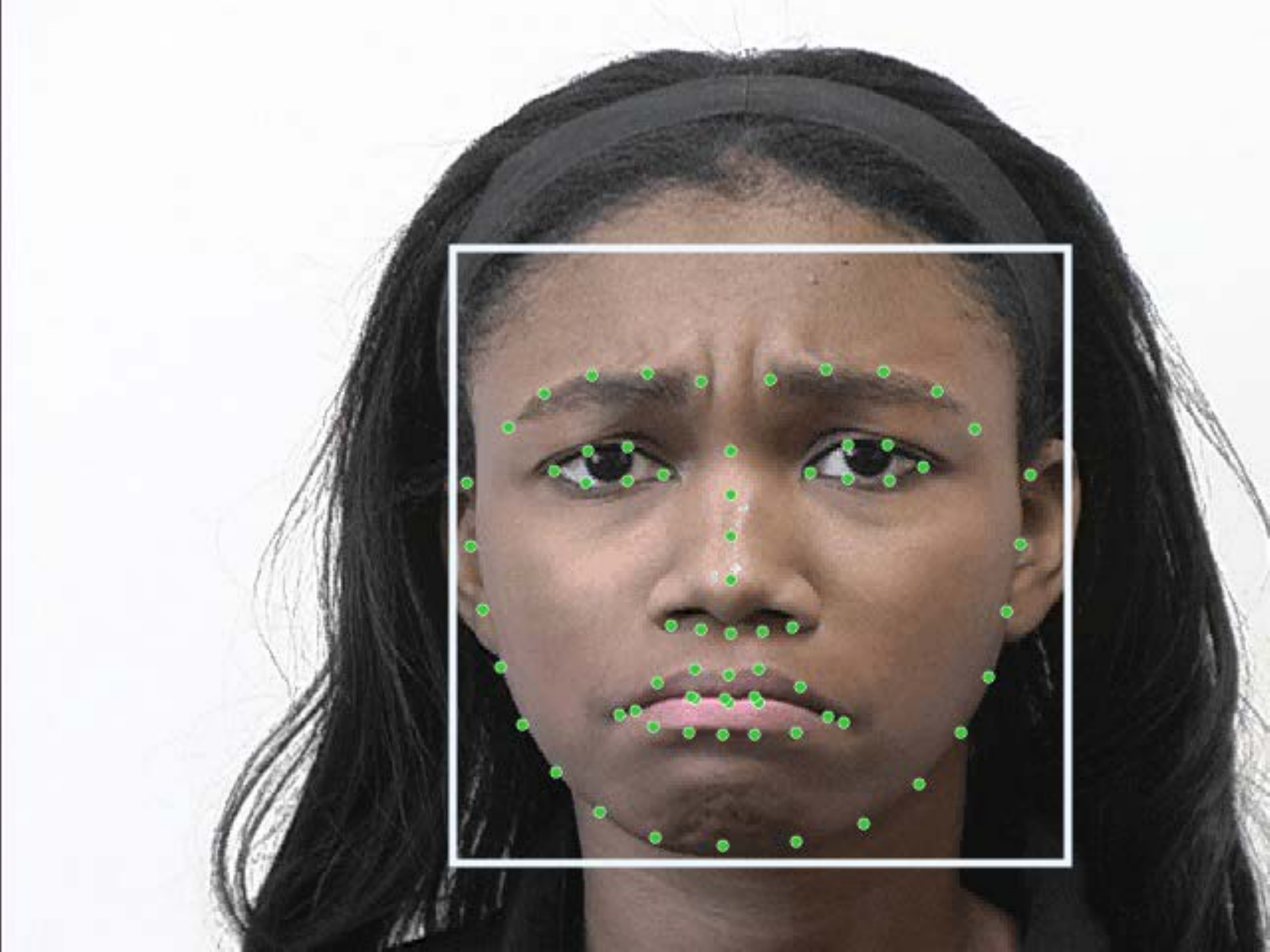}}}
	
	\caption{Rectangles enclose the face regions, green markings indicate the facial landmarks.
		Landmarks move distinctively with different facial expressions.}
	\label{fig:Cohn-Kanade}
\end{figure}

We start by detecting the face region with the Viola and Jones algorithm~\cite{Viola:2004}.
Then, 68 facial landmarks are localized using Kazemi and Sullivan's method~\cite{Kazemi:2014}.
See Figure~\ref{fig:Cohn-Kanade} for face detections and located landmarks on neutral and expressive face images taken from the CK+ dataset.
The distances between landmark pairs change as the subject expresses an emotion.
We use the horizontal and vertical variations in these distances as features.
A descriptive subset is chosen among these features using forward sequential selection, and used for classification by an SVM.

\subsection{Extracting Spatial Features}
\label{sec:features}

In the feature extraction step, the horizontal and vertical distances between all landmark pairs are calculated.
Using the relative displacements of landmarks provides robustness against translations between the neutral and the expressive face.
By taking the difference between distance vectors obtained from a neutral and an expressive face, the displacement caused by the facial expression is captured~\cite{Lucey:2010}.
This approach can also be interpreted as an implicit calibration using the neutral state of the face.

The CK+ dataset provides a set of consecutive frames where the subject gradually displays the intended expression.
While extracting features, we only use the first and the last frames, which are fully neutral and fully expressive (see Figure~\ref{fig:Cohn-Kanade}).
For both images, 68 facial landmarks are located, which form 2278 different pairs.
Since we handle the horizontal and vertical distances between the landmarks independently, a distance vector with the size of 4556 is obtained from each image in the pair.
The difference of these two distance vectors results in a feature vector of the same size for each example.
This large feature vector includes non-descriptive and redundant elements.

\subsection{Sequential Forward Selection of Features}
\label{sec:sfs}

In Section~\ref{sec:features}, we extracted a large feature vector, composed of non-descriptive and redundant features, along with useful ones.
To form a descriptive subset, we use sequential forward selection~(SFS).
This is a greedy search algorithm that iteratively selects the feature that improves the recognition accuracy the most.
A feature's usefulness is defined by the improvement it provides to recognition accuracy when used with the previously selected features.

Before starting SFS, we randomly segment CK+ dataset as the training and test set.
At the start of the $k^\text{th}$ iteration of the algorithm, $k-1$ features are already selected.
Features that are not among the selected are candidates.
To test a candidate, it is grouped with the selected features, and the resulting vector is $L2$ normalized.
The normalized feature vectors from the training set are used to train a multiclass SVM.
This classifier uses the normalized vectors from the test set for recognition.
The candidate whose addition improves the recognition accuracy the most is selected and the algorithm moves on to the next iteration.
The algorithm stops when none of the candidates can improve recognition accuracy.

\section{Experimental Results}

\begin{table}
	\centering
	\caption{Number of examples for each facial expression in the CK+ dataset.}
	\tabulinesep=1mm
	\vspace{7mm}
	\newcommand{\tablecolumnsize}{0.075\columnwidth}
	
	\begin{tabu} {
			>{\centering}m{\tablecolumnsize}
			>{\centering}m{\tablecolumnsize}
			>{\centering}m{\tablecolumnsize}
			>{\centering}m{\tablecolumnsize}
			>{\centering}m{\tablecolumnsize}
			>{\centering}m{\tablecolumnsize}
			>{\centering}m{\tablecolumnsize}
			>{\centering}m{\tablecolumnsize}
		}
		\begin{rotate}{45} Anger \end{rotate} &
		\begin{rotate}{45} Contempt \end{rotate} &
		\begin{rotate}{45} Disgust \end{rotate} &
		\begin{rotate}{45} Fear \end{rotate} &
		\begin{rotate}{45} Happiness \end{rotate} &
		\begin{rotate}{45} Sadness \end{rotate} &
		\begin{rotate}{45} Surprise \end{rotate} &
		\begin{rotate}{45} \textbf{Total} \end{rotate} \\
		\tabucline[1.5pt]{-}
		45 & 19 & 59 & 25 & 69 & 28 & 82 & \textbf{327}
	\end{tabu}
	\label{tab:classes}
\end{table}

We conducted our experiments on the extended Cohn-Kanade dataset~\cite{Lucey:2010}.
The CK+ dataset contains images of faces with seven different facial expressions.
These expressions are labeled as anger, contempt, disgust, fear, happiness, sadness and surprise.
The dataset is composed of 327 image sequences gathered from 123 subjects.
These image sets are consecutive frames that start at a neutral expression and end when the subject is expressing the respective emotion intensely.
Number of examples for each expression is given in Table~\ref{tab:classes}\footnote{
	The Contempt example from Subject-129 is mislabeled as Surprised in the dataset.
	We used the corrected label.}.
Since gathering definitive examples from classes such as contempt and fear is more difficult, examples from these classes are lower in number.

\subsection{Feature Selection}

\begin{figure}
	\captionsetup[subfigure]{labelformat=empty}
	\centering	
	\subfloat{{\includegraphics[width = 0.2\textwidth]{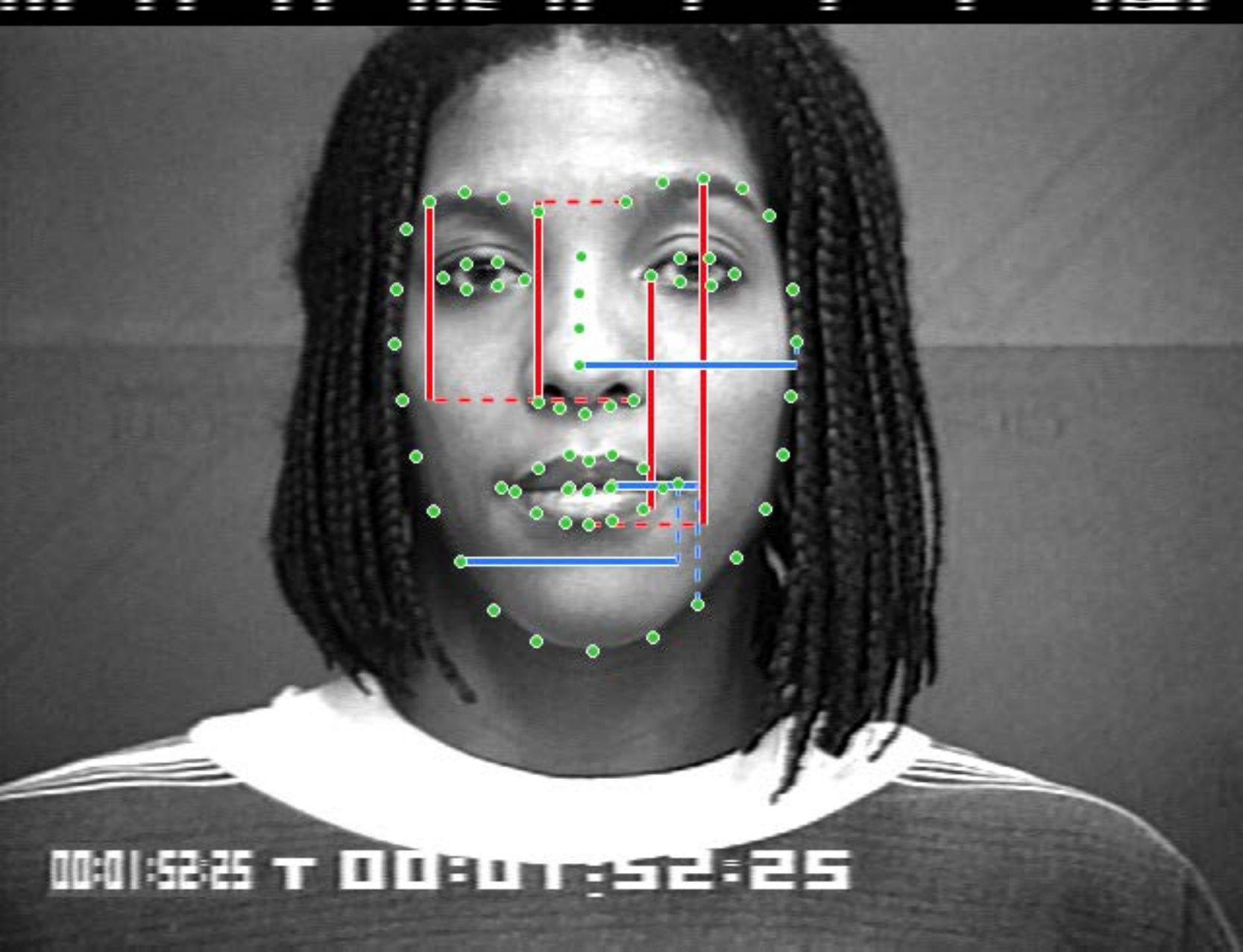}}}
	\subfloat{{\includegraphics[width = 0.2\textwidth]{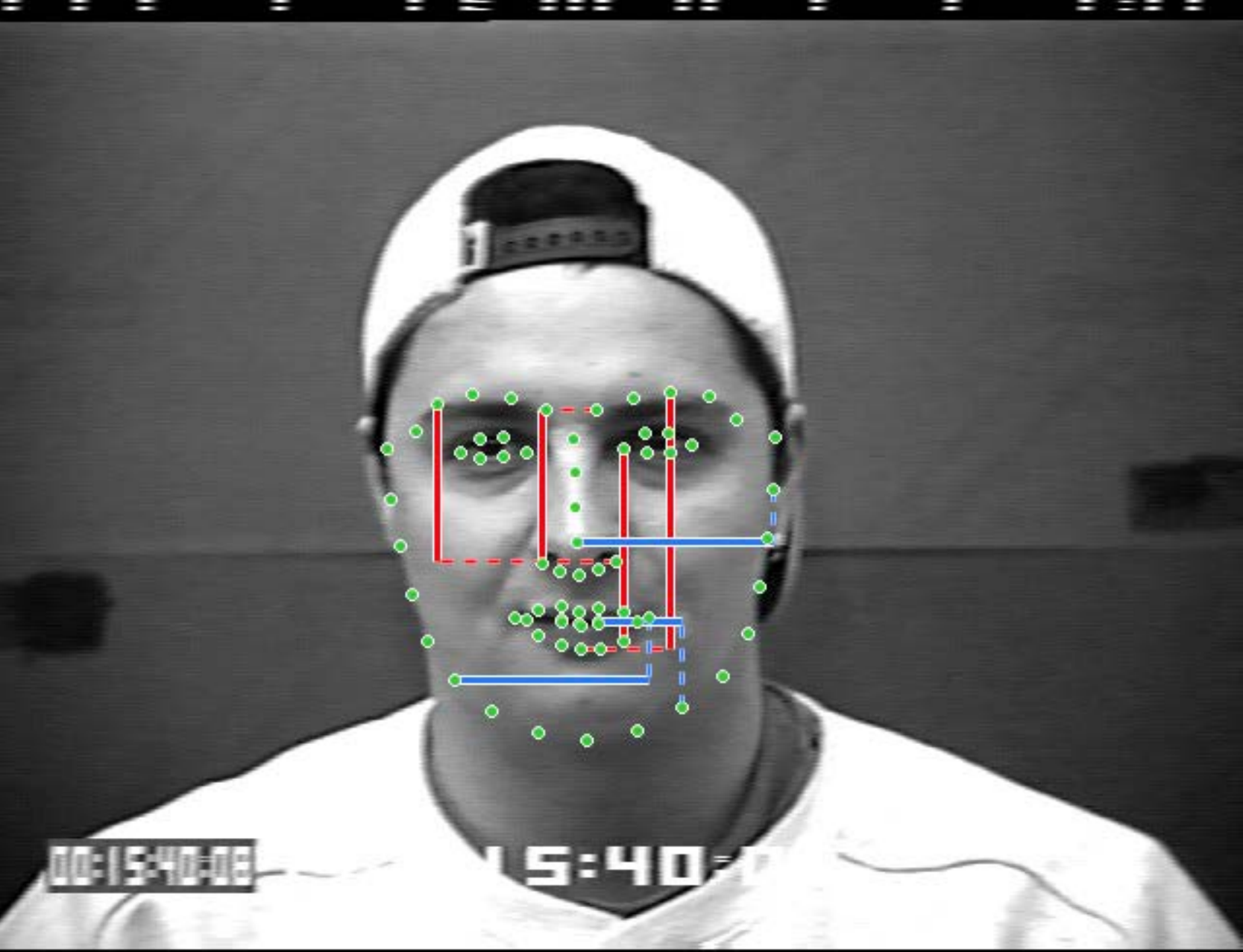}}}
	\subfloat{{\includegraphics[width = 0.2\textwidth]{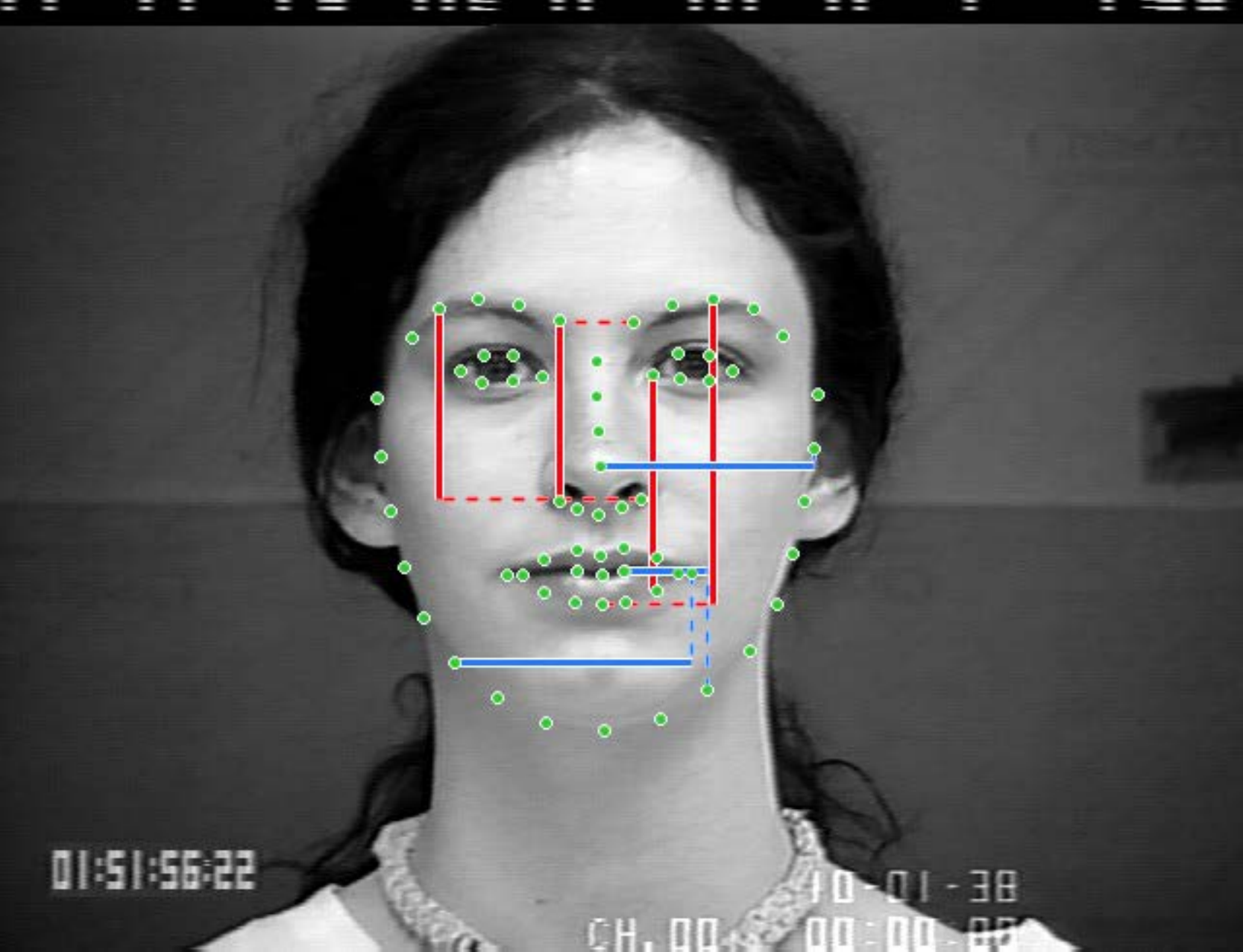}}}
	\subfloat{{\includegraphics[width = 0.2\textwidth]{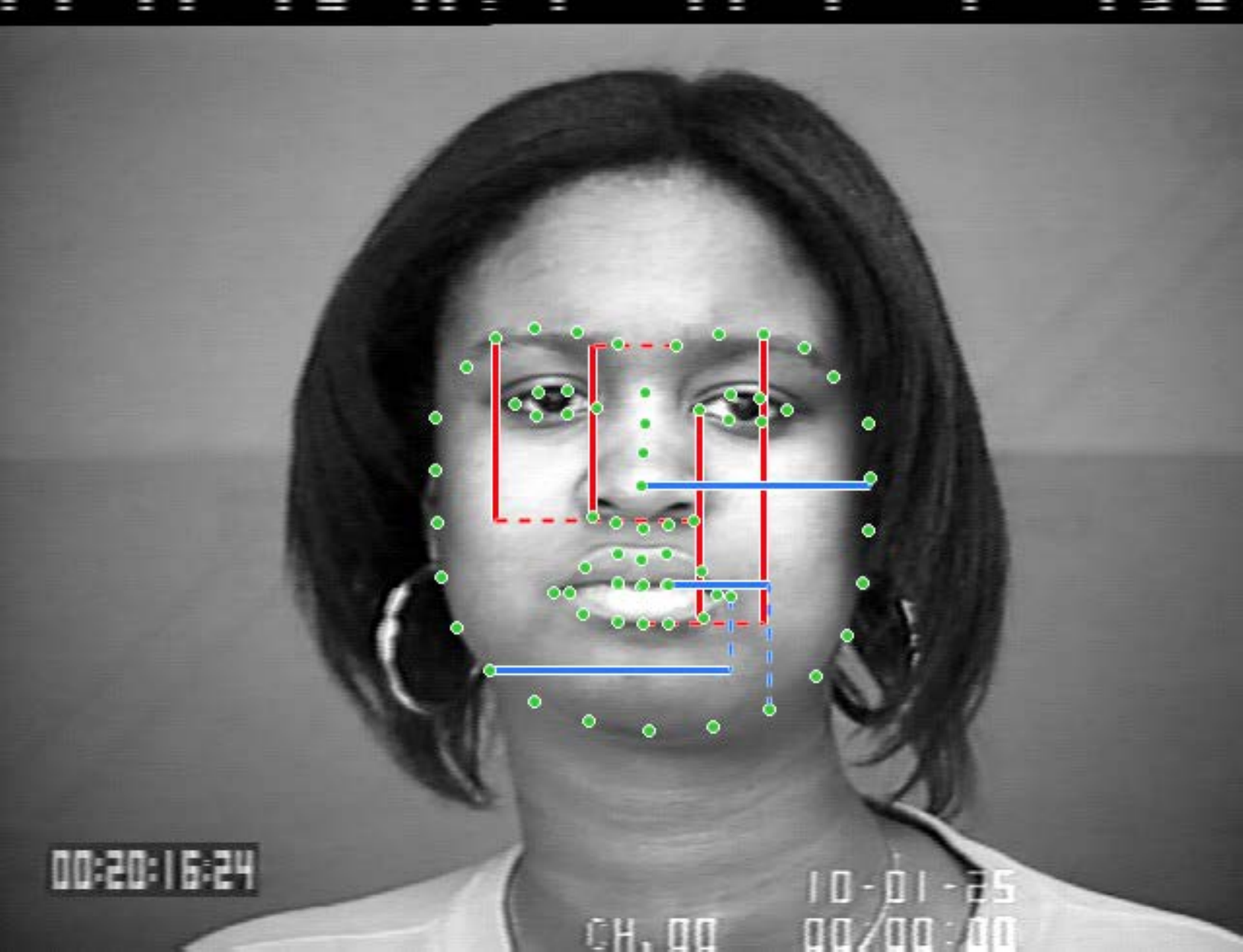}}}
	\subfloat{{\includegraphics[width = 0.2\textwidth]{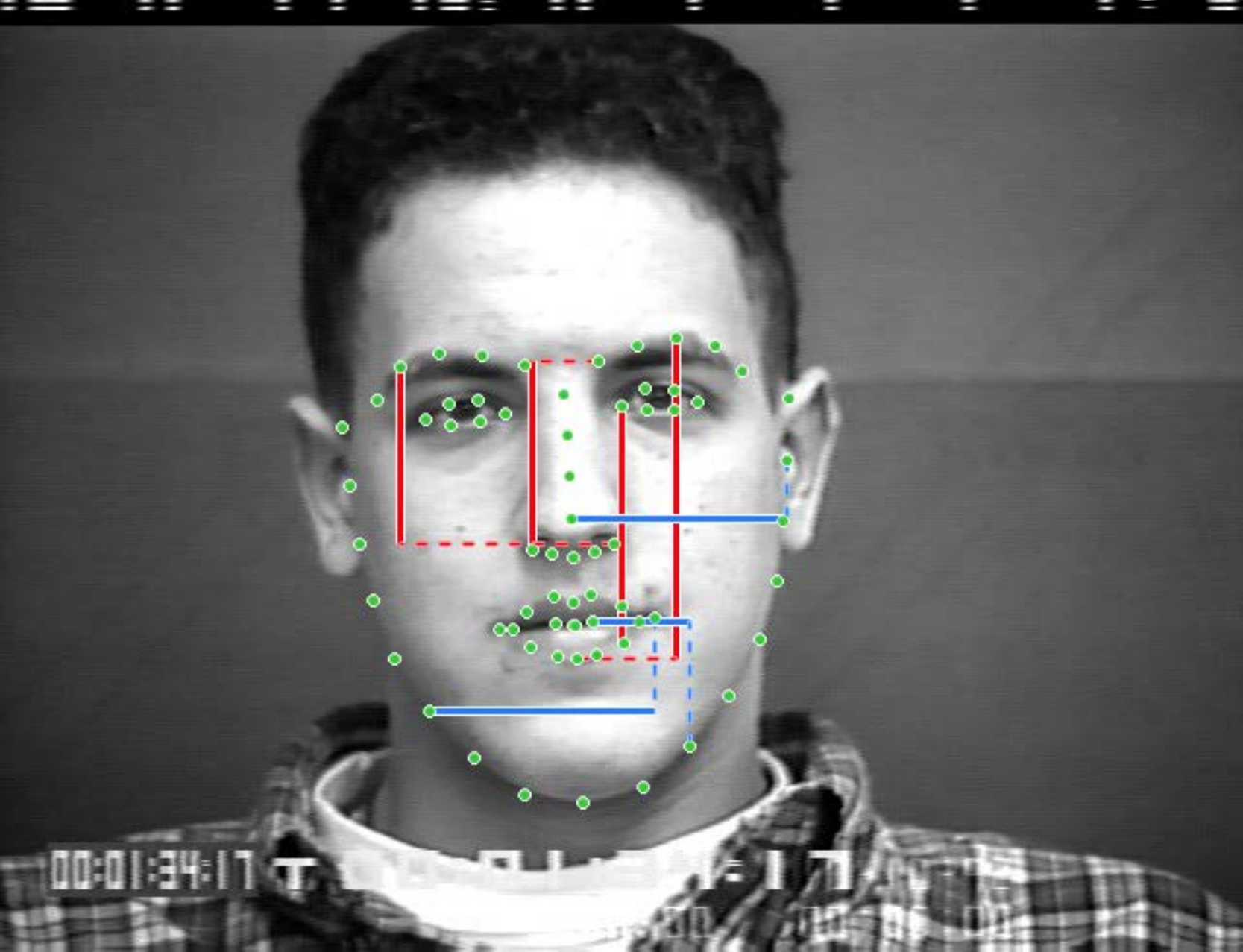}}}
	\vspace{-3mm}
	\subfloat[Anger]{{\includegraphics[width = 0.2\textwidth]{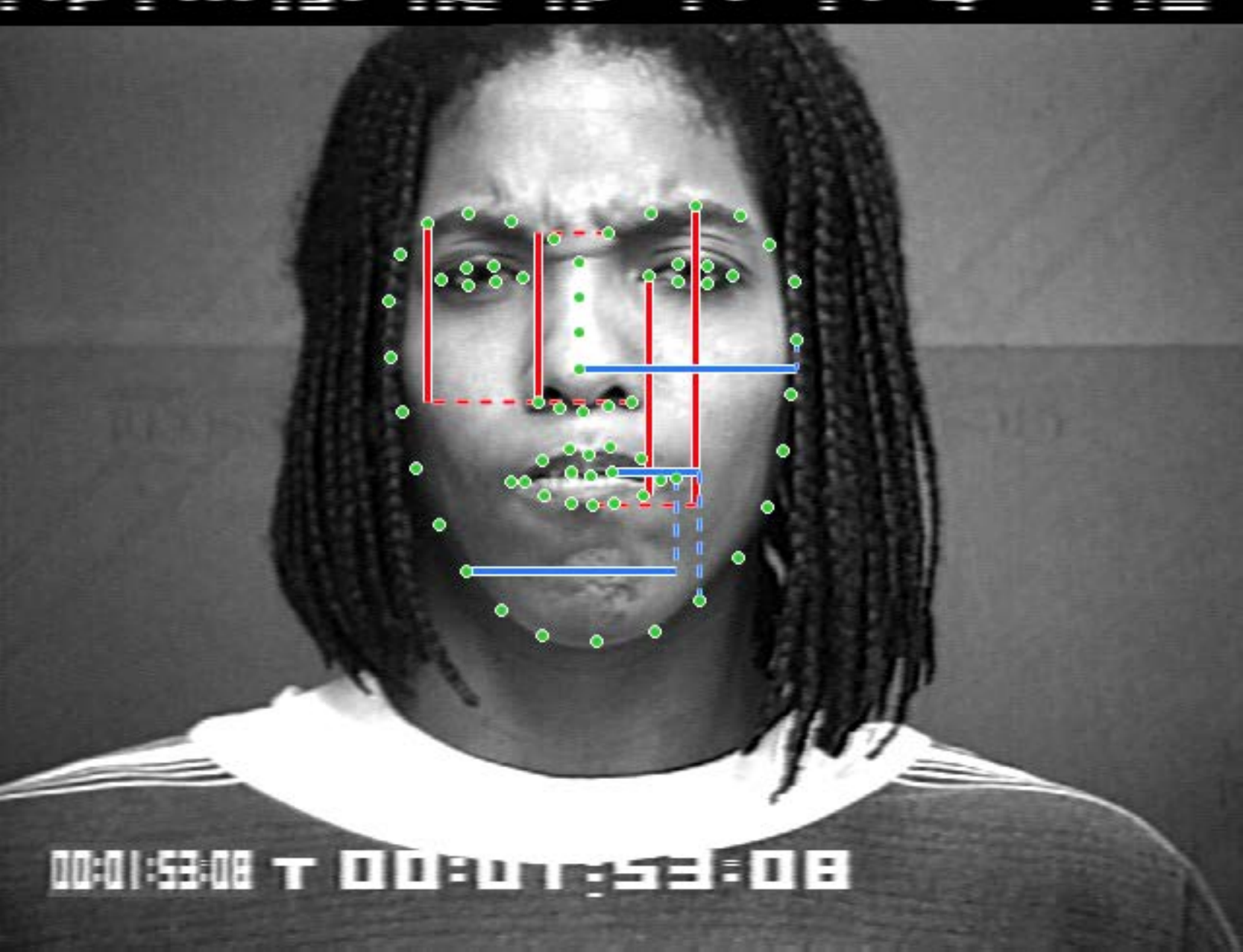}}}
	\subfloat[Disgust]{{\includegraphics[width = 0.2\textwidth]{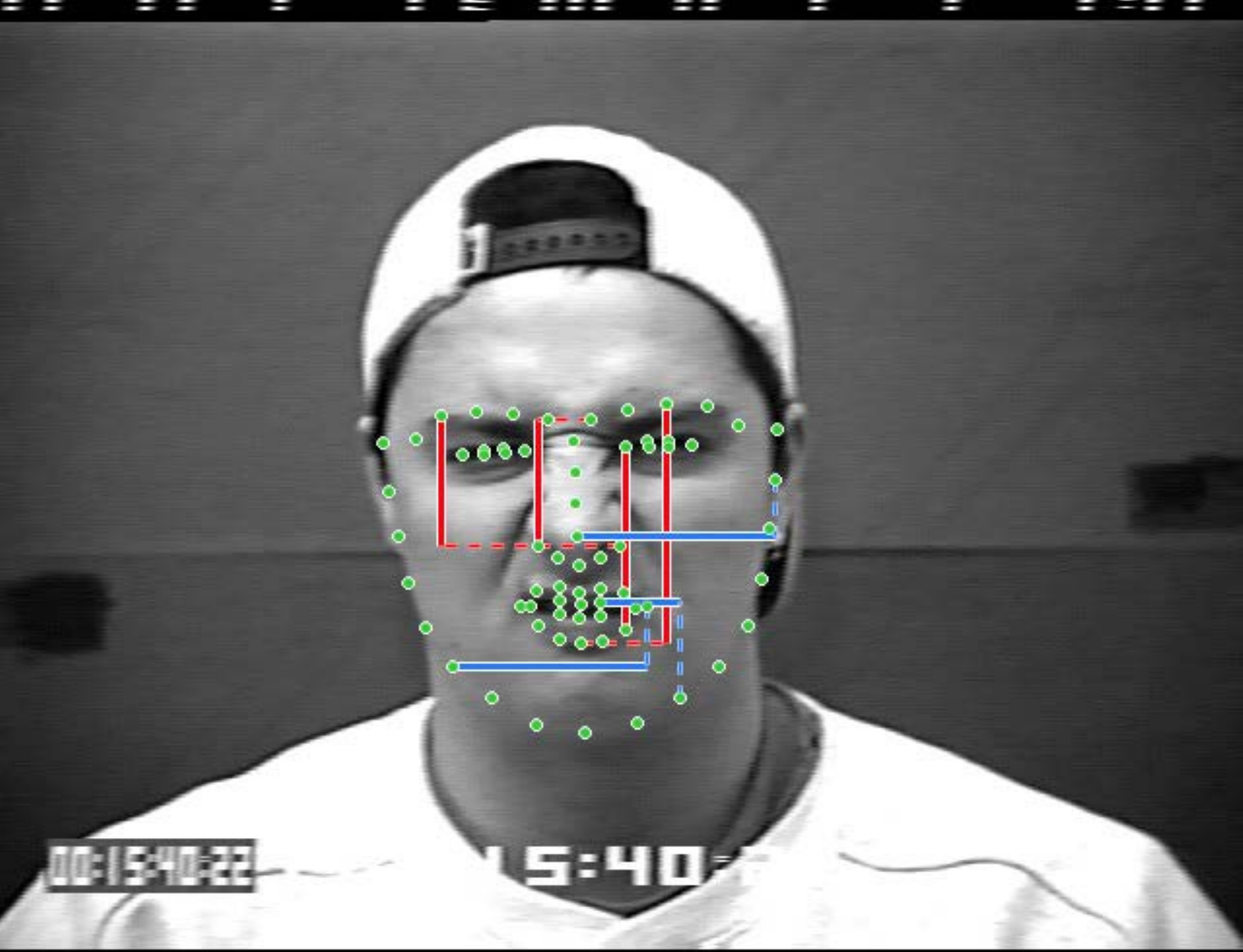}}}
	\subfloat[Happiness]{{\includegraphics[width = 0.2\textwidth]{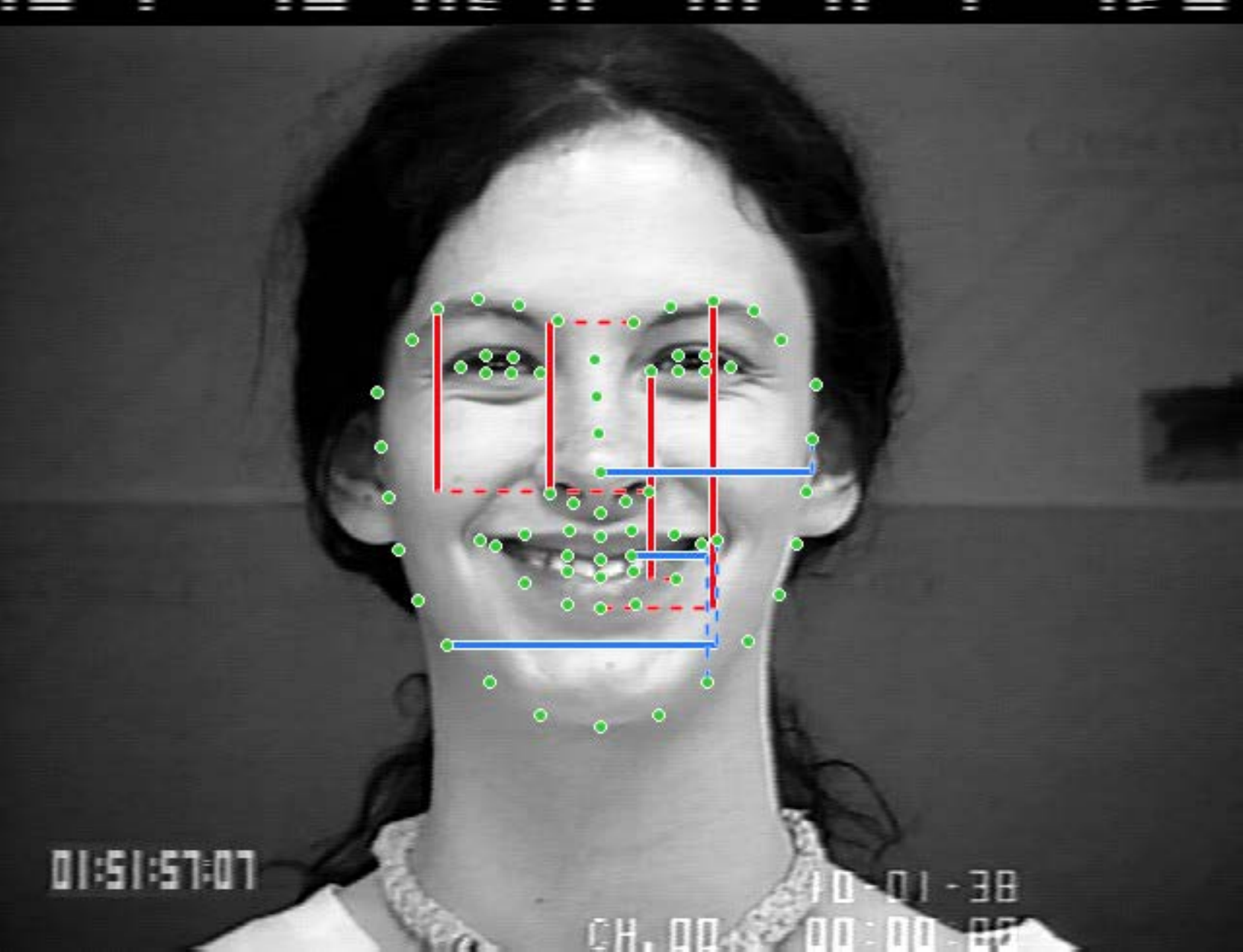}}}
	\subfloat[Sadness]{{\includegraphics[width = 0.2\textwidth]{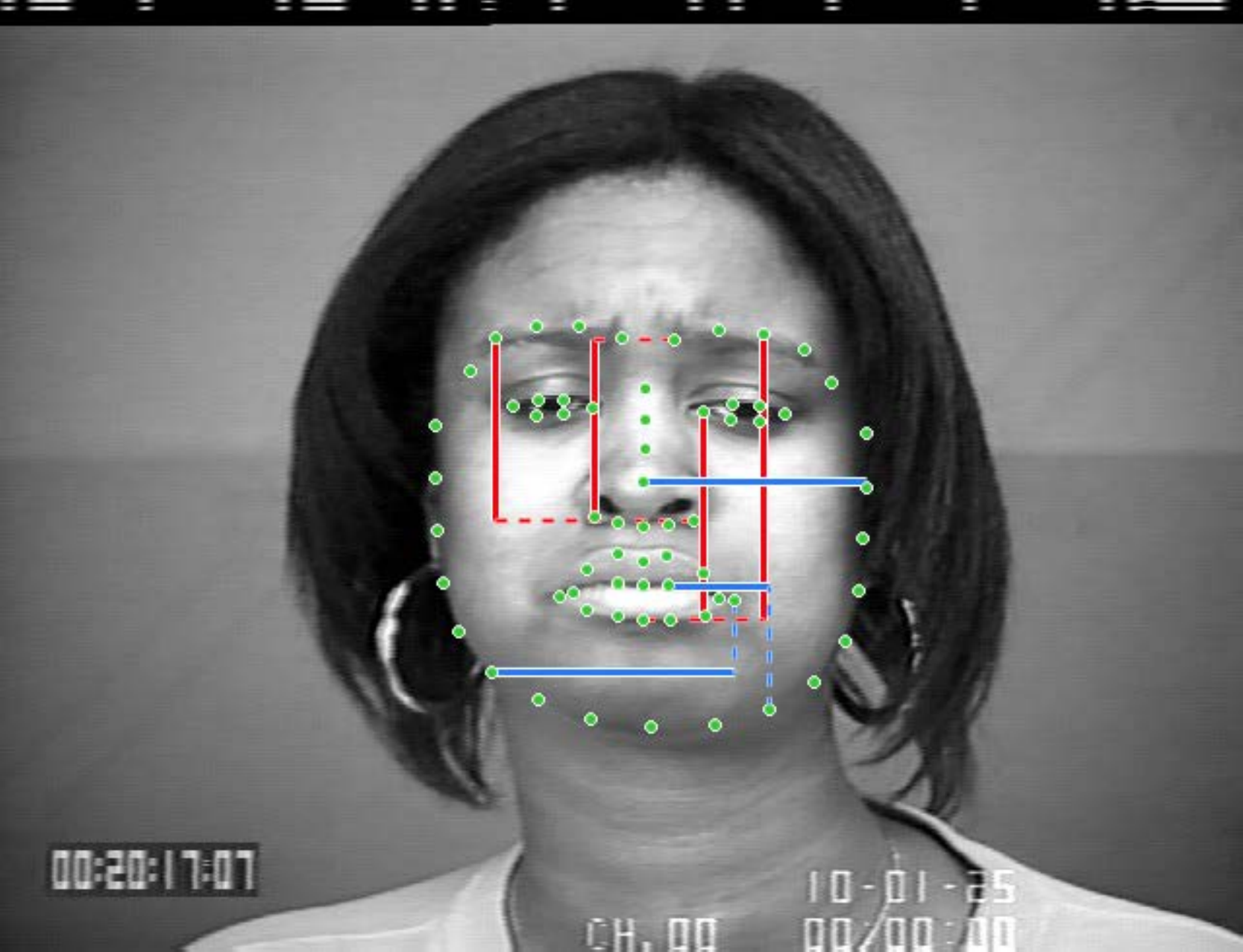}}}		
	\subfloat[Surprise]{{\includegraphics[width = 0.2\textwidth]{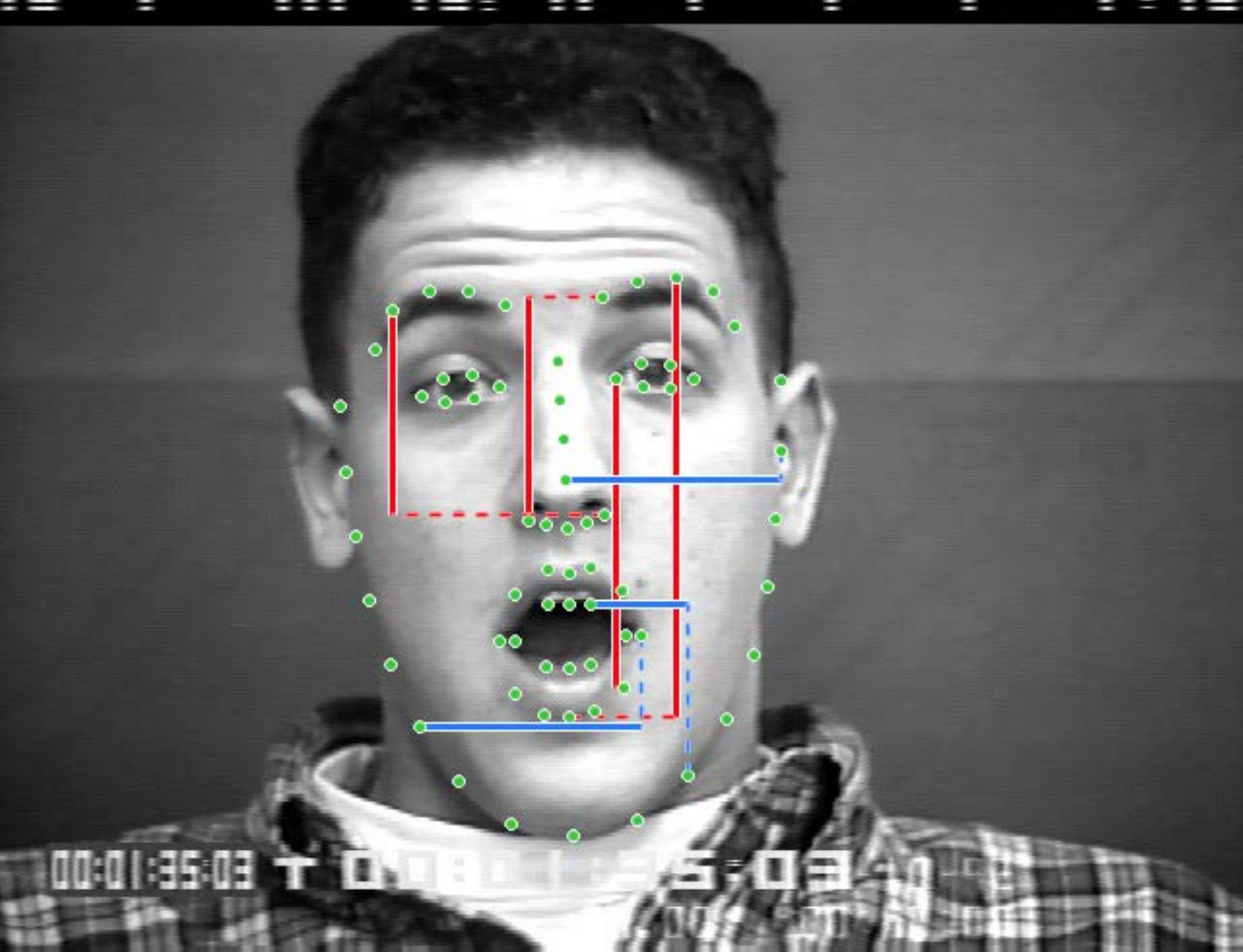}}}
	\caption{
		The blue bars show the horizontal and the red bars show the vertical distances between two facial landmarks.
		Each bar starts at the first facial landmark, and its end is connected to the second facial landmark with a dashed line.
		The changes in the lengths of these bars are the features used for classification.
		Note that for each expression, at least one of the bars shorten or elongate distinctly.}
	\label{fig:Features}
	\vspace{3mm}
\end{figure}

To apply SFS as described in Section~\ref{sec:sfs}, the dataset is segmented into the training and test sets with $0.6$ and $0.4$ ratios, while protecting the class frequencies of the original dataset.
The features chosen by SFS are illustrated in Figure~\ref{fig:Features}.
The distances used as features are plotted as horizontal or vertical bars.
Lengths of these bars change distinctly with the respective expression.

Although these features are chosen automatically, they are justifiable when inspected individually.
It can be said that the movements of the mouth and eyebrows are particularly effective in recognizing facial expressions.
Three of the features describe the movements of the eyebrows, while four describe movements of the mouth.
The bottommost blue bar only describes the widening of the mouth, as its other end is anchored to a stationary part of the jaw~(see Figure~\ref{fig:Features}, Happiness).
The leftmost red bar is another feature that describes a single factor, the vertical eyebrow movement~(see Figure~\ref{fig:Features}, Disgust and Surprise).

An unexpected feature is the horizontal distance variation between the ear and the nose.
SFS chooses features even when they provide a very marginal increase in accuracy.
Assuming this was the case, we tried eliminating this feature, which resulted in a 3\% decrease in accuracy.
Considering that additional features improve classification accuracy with diminishing returns, this difference is actually significant.
We speculate that the role of this feature may be to sense head pose variation.
Since the distance from a person's nose to ear cannot change, the feature extracted from this landmark pair will be non-zero only when the head pose changes.
Head movement is limited to intense expressions such as anger and surprise, which may be the reason that this feature is descriptive.

\subsection{Classification Accuracy}

\begin{table}
	\caption{Confusion matrix of classification using the selected spatial features in the CK+ dataset.}
	\vspace{5mm}
	
	\newcommand\gray{gray}
	
	\newcommand\ColCell[1]{%
		\pgfmathparse{#1<.5?1:0}%
		\ifnum\pgfmathresult=0\relax\color{white}\fi
		\pgfmathparse{1-#1}%
		\expandafter\cellcolor\expandafter[%
		\expandafter\gray\expandafter]\expandafter{\pgfmathresult}#1}
	
	\newcolumntype{E}{>{\collectcell\ColCell}c<{\endcollectcell}}
	\centering
	\resizebox{0.75\columnwidth}{!}
	{
		\begin{tabular}{c*{7}{|E}|}
			\multicolumn{1}{c}{} & \multicolumn{1}{c}{\begin{rotate}{45} Anger \end{rotate}} & \multicolumn{1}{c}{\begin{rotate}{45} Contempt \end{rotate}} 
			& \multicolumn{1}{c}{\begin{rotate}{45} Disgust \end{rotate}} & \multicolumn{1}{c}{\begin{rotate}{45} Fear \end{rotate}} & \multicolumn{1}{c}{\begin{rotate}{45} Happiness \end{rotate}} 
			& \multicolumn{1}{c}{\begin{rotate}{45} Sadness \end{rotate}} & \multicolumn{1}{c}{\begin{rotate}{45} Surprise \end{rotate}} \\ \hhline{~*7{|-}|}
			Anger & 0.78 & 0.04 & 0.09 & 0 & 0 & 0.09 & 0 \\ \hhline{~*7{|-}|}
			Contempt & 0.21 & 0.64 & 0 & 0.05 & 0.05 & 0.05 & 0 \\ \hhline{~*7{|-}|} 
			Disgust & 0.05 & 0 & 0.93 & 0.02 & 0 & 0 & 0 \\ \hhline{~*7{|-}|}
			Fear & 0 & 0 & 0 & 0.80 & 0.04 & 0 & 0.16 \\ \hhline{~*7{|-}|}
			Happiness & 0 & 0 & 0 & 0.01 & 0.99 & 0 & 0 \\ \hhline{~*7{|-}|}
			Sadness & 0.11 & 0.11 & 0.03 & 0.03 & 0 & 0.64 & 0.08 \\ \hhline{~*7{|-}|}
			Surprise & 0 & 0 & 0 & 0 & 0 & 0 & 1 \\ \hhline{~*7{|-}|}
		\end{tabular}
	}
	
	\label{tab:confmat}
\end{table}

\begin{figure}
	\captionsetup[subfigure]{labelformat=empty,justification=centering}
	\centering
	\subfloat[Surprise - 94\%]{{\includegraphics[width=0.25\columnwidth]{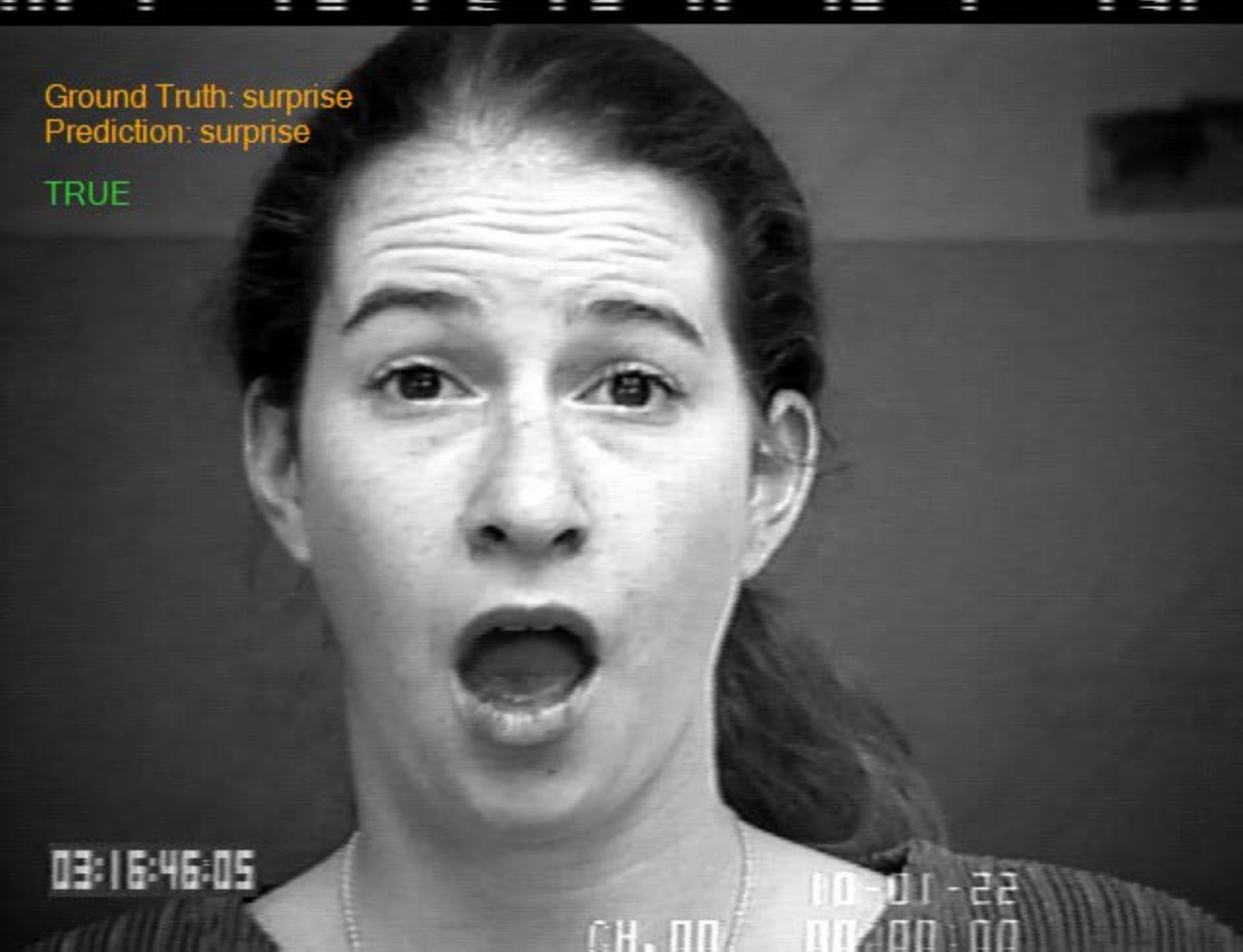}}}
	\subfloat[Anger - 92\%]{{\includegraphics[width=0.25\columnwidth]{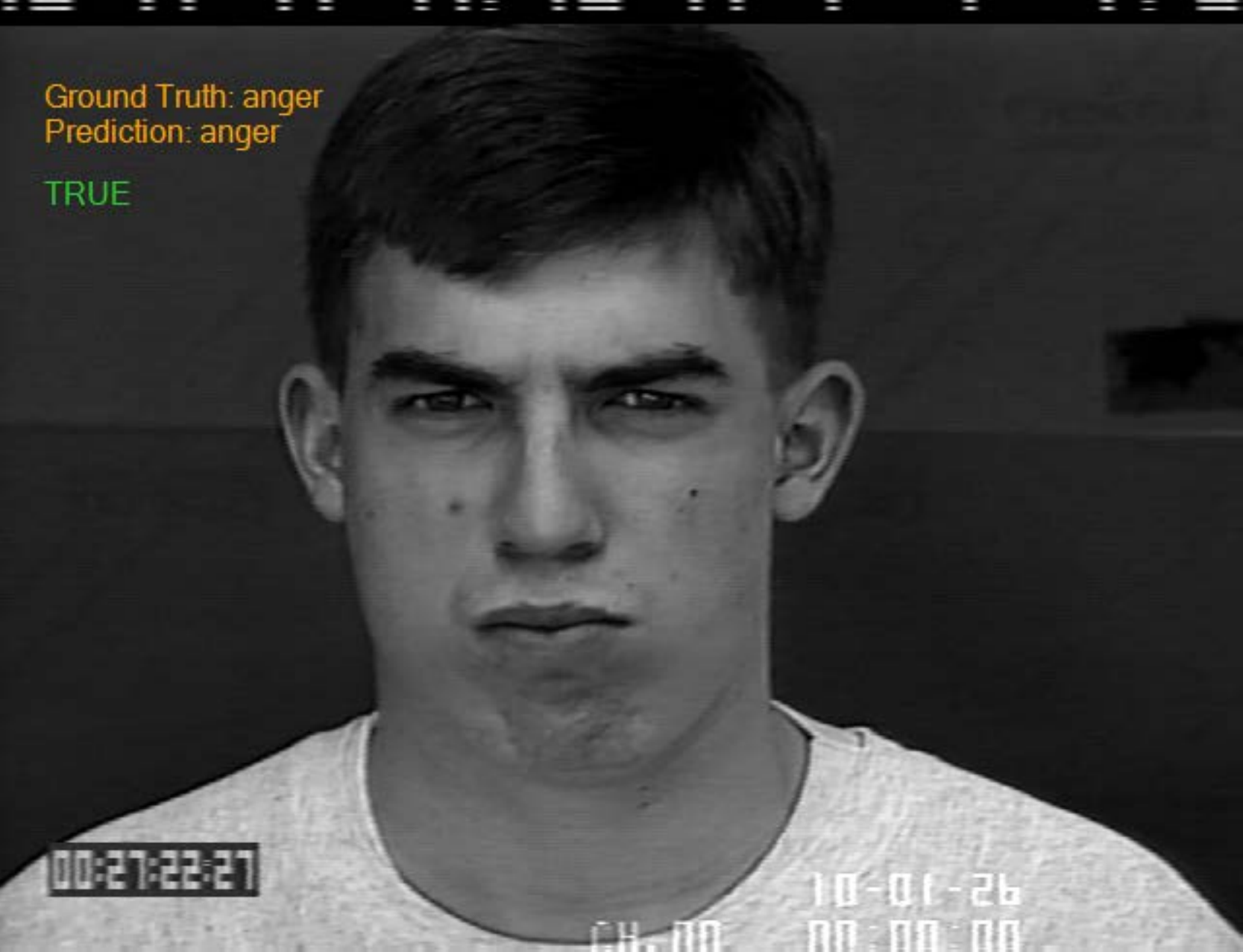}}}
	\subfloat[Fear - 97\%]{{\includegraphics[width=0.25\columnwidth]{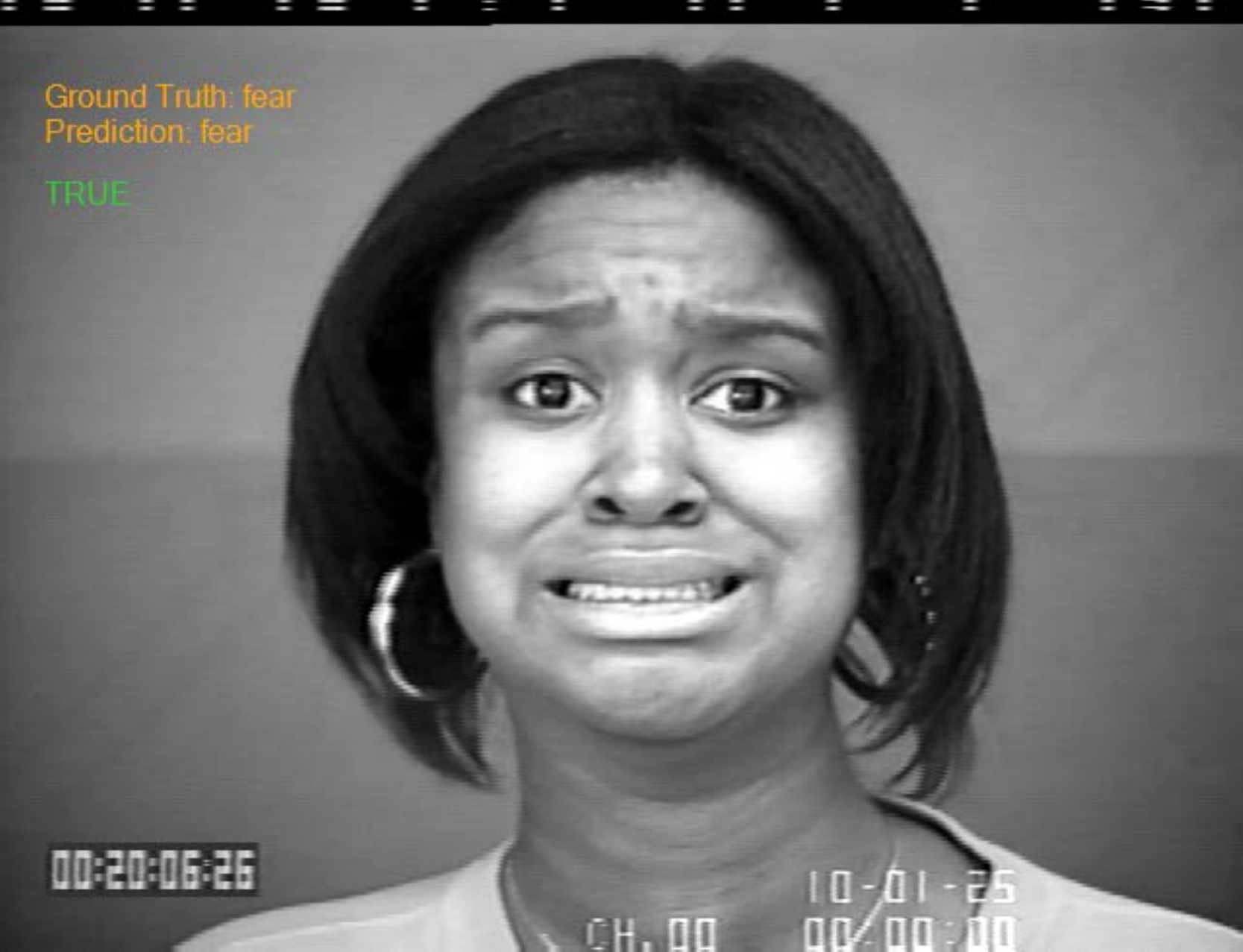}}}
	\subfloat[Sadness - 98\%]{{\includegraphics[width=0.25\columnwidth]{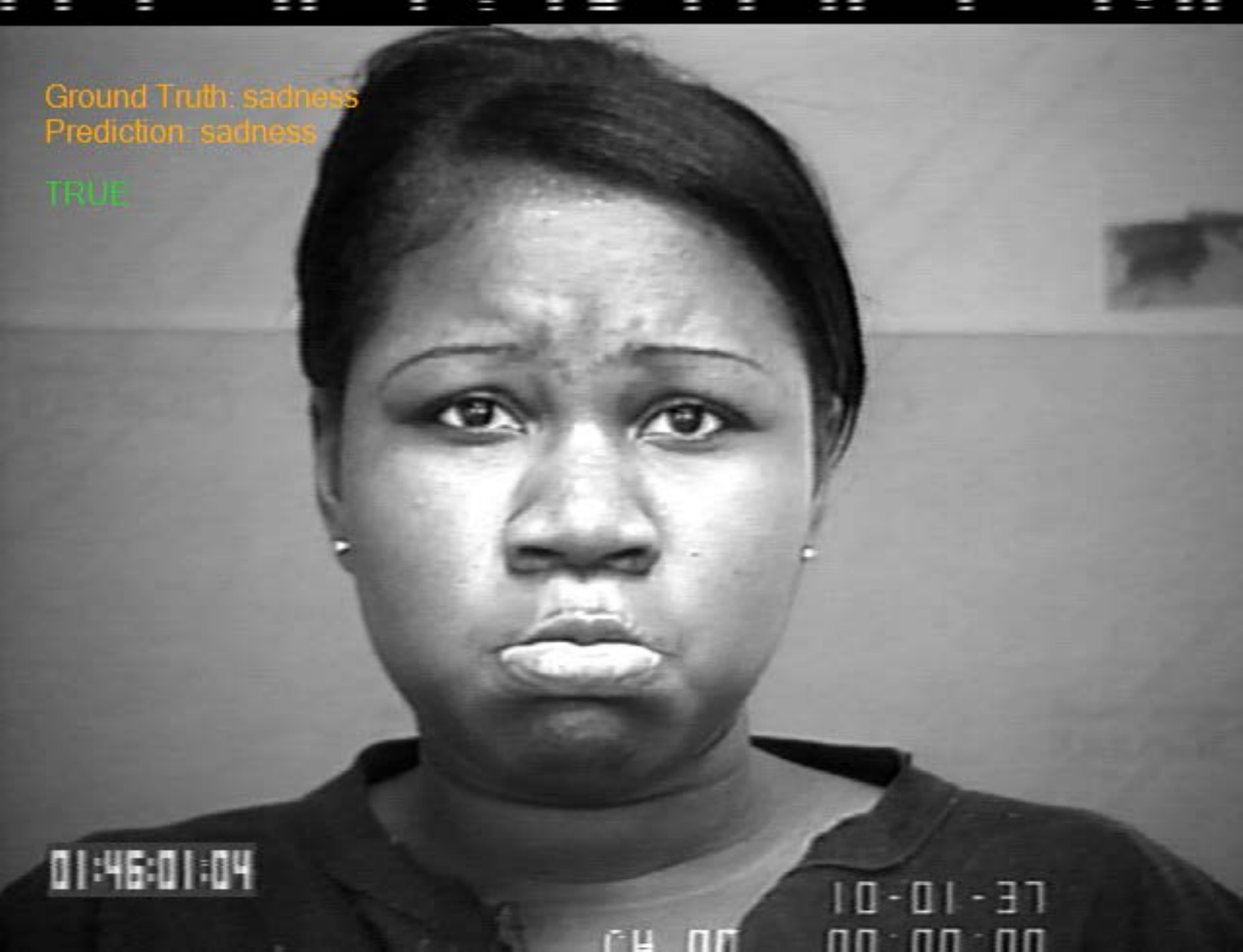}}}
	
	\subfloat[Contempt - 81\%]{{\includegraphics[width=0.25\columnwidth]{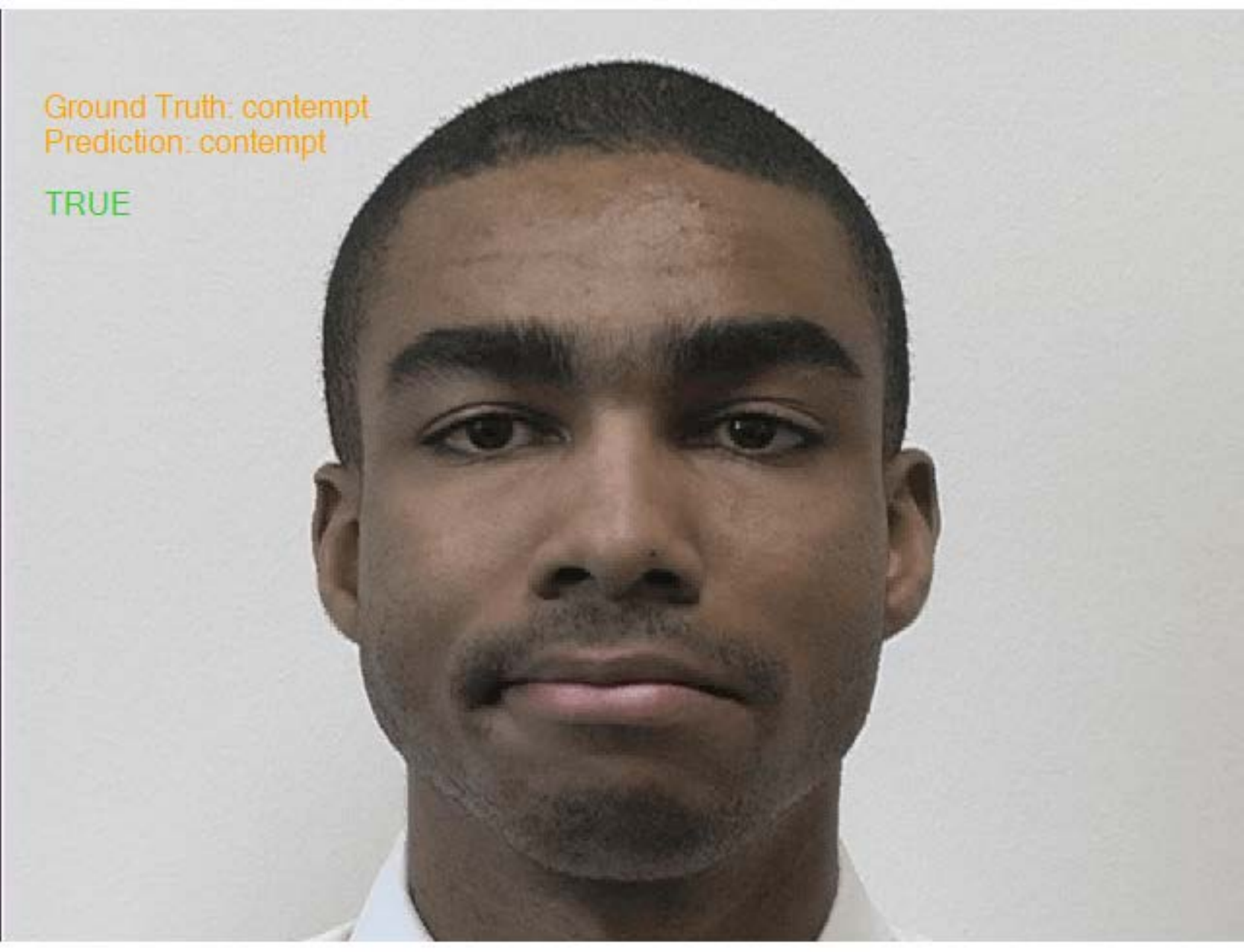}}}
	\subfloat[Happiness - 99\%]{{\includegraphics[width=0.25\columnwidth]{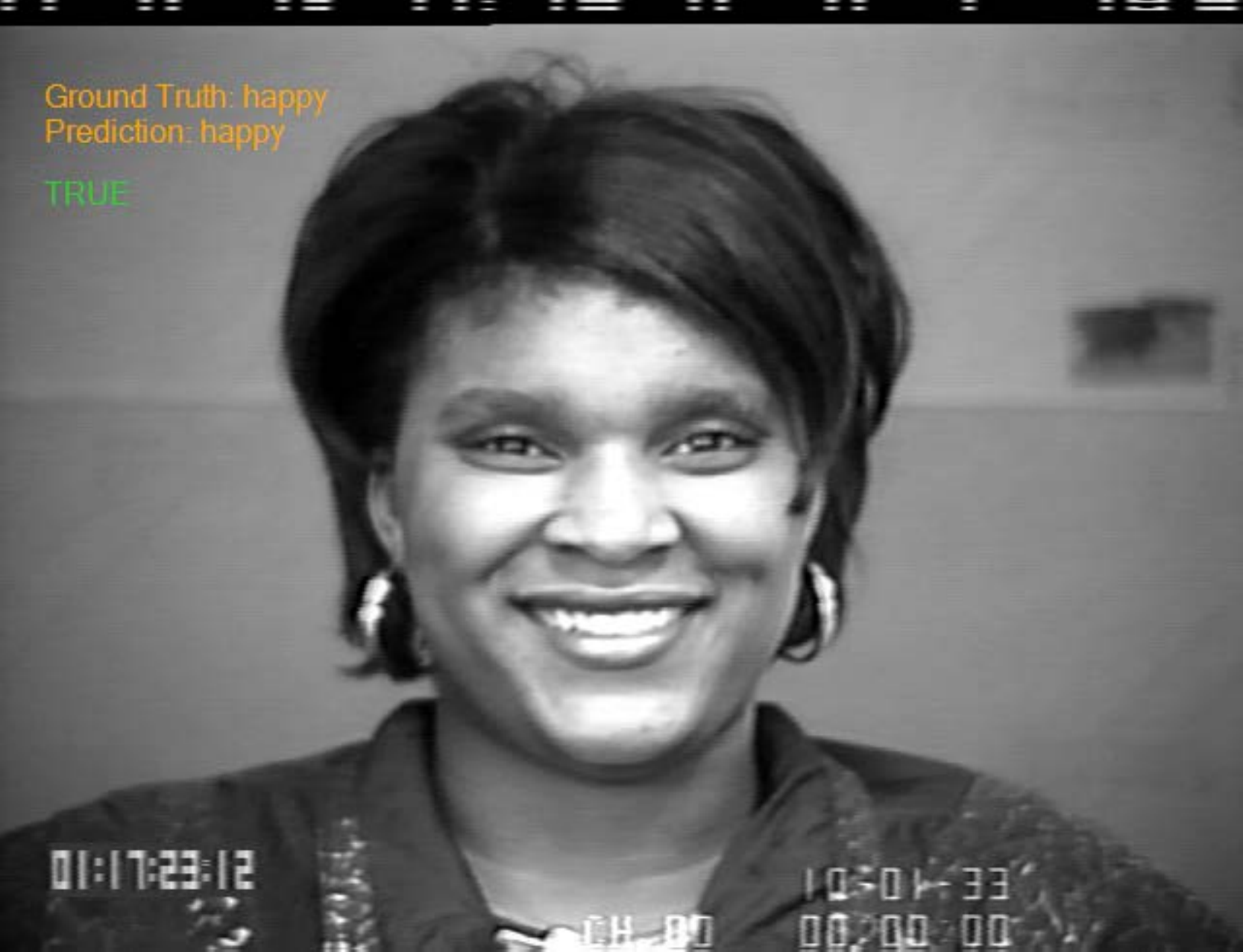}}}
	\subfloat[Disgust - 88\%]{{\includegraphics[width=0.25\columnwidth]{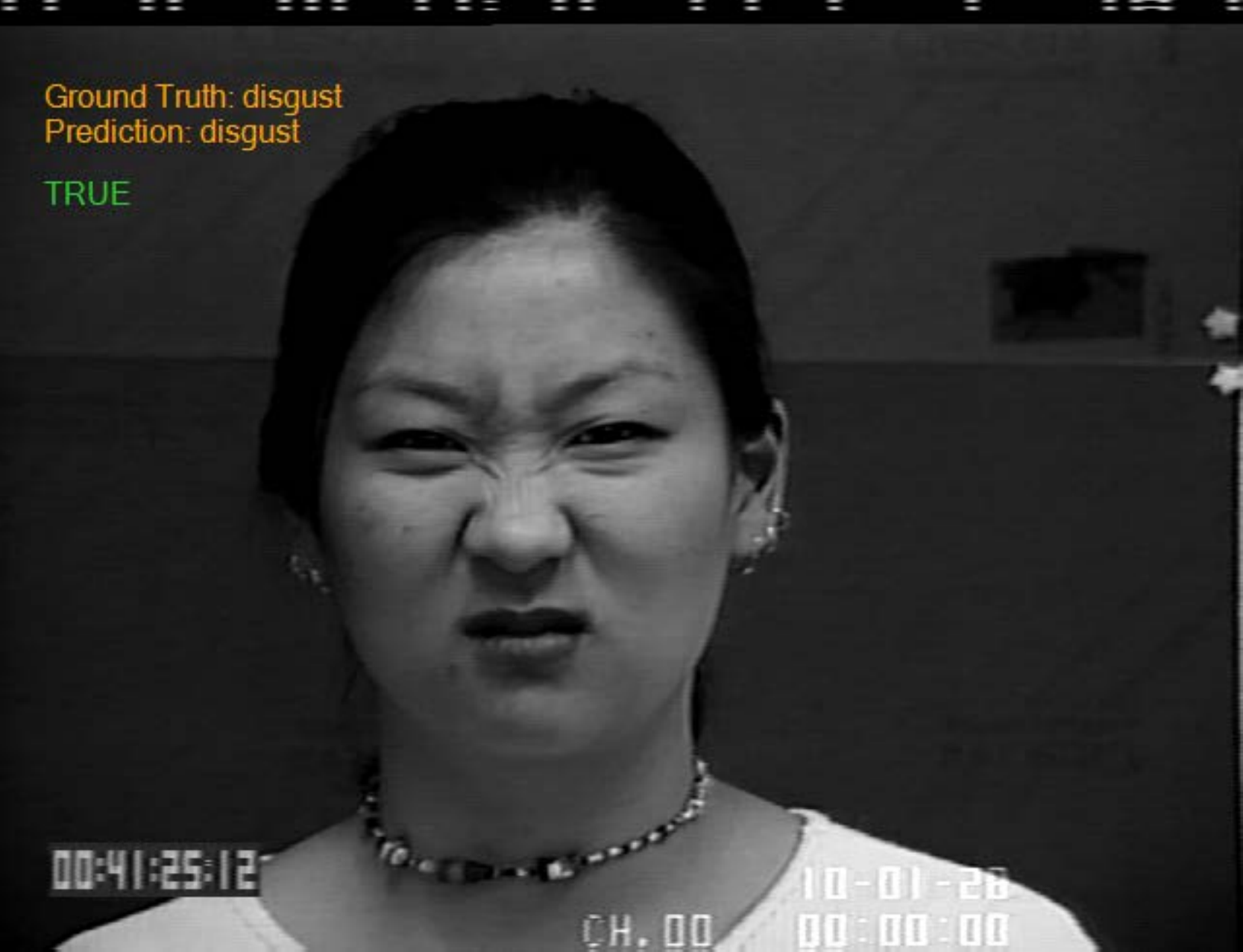}}}
	\subfloat[Surprise - 92\%]{{\includegraphics[width=0.25\columnwidth]{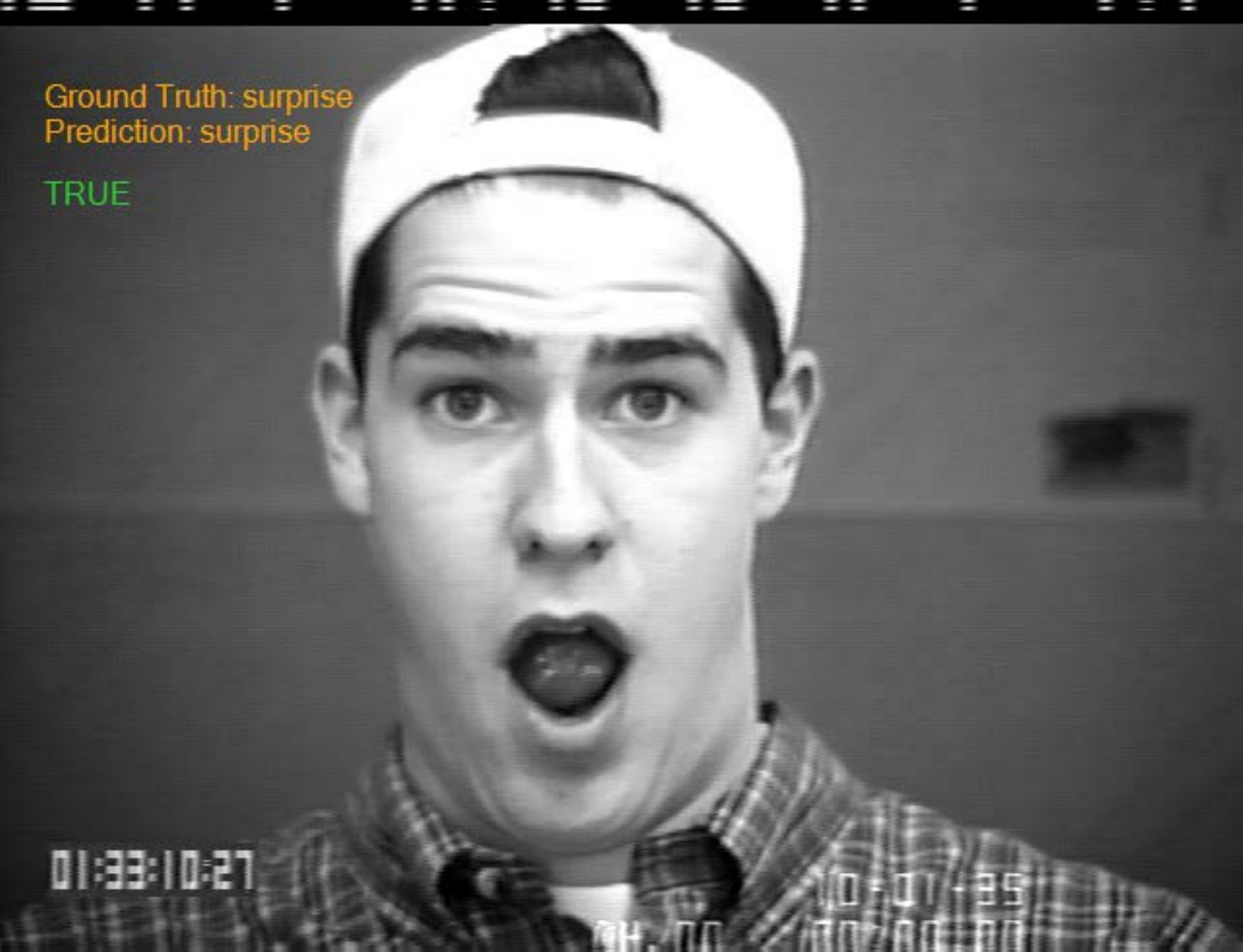}}}
	
	\subfloat[][Sadness - 91\% \\ GT: Contempt - 0\%]{{\includegraphics[width=0.25\columnwidth]{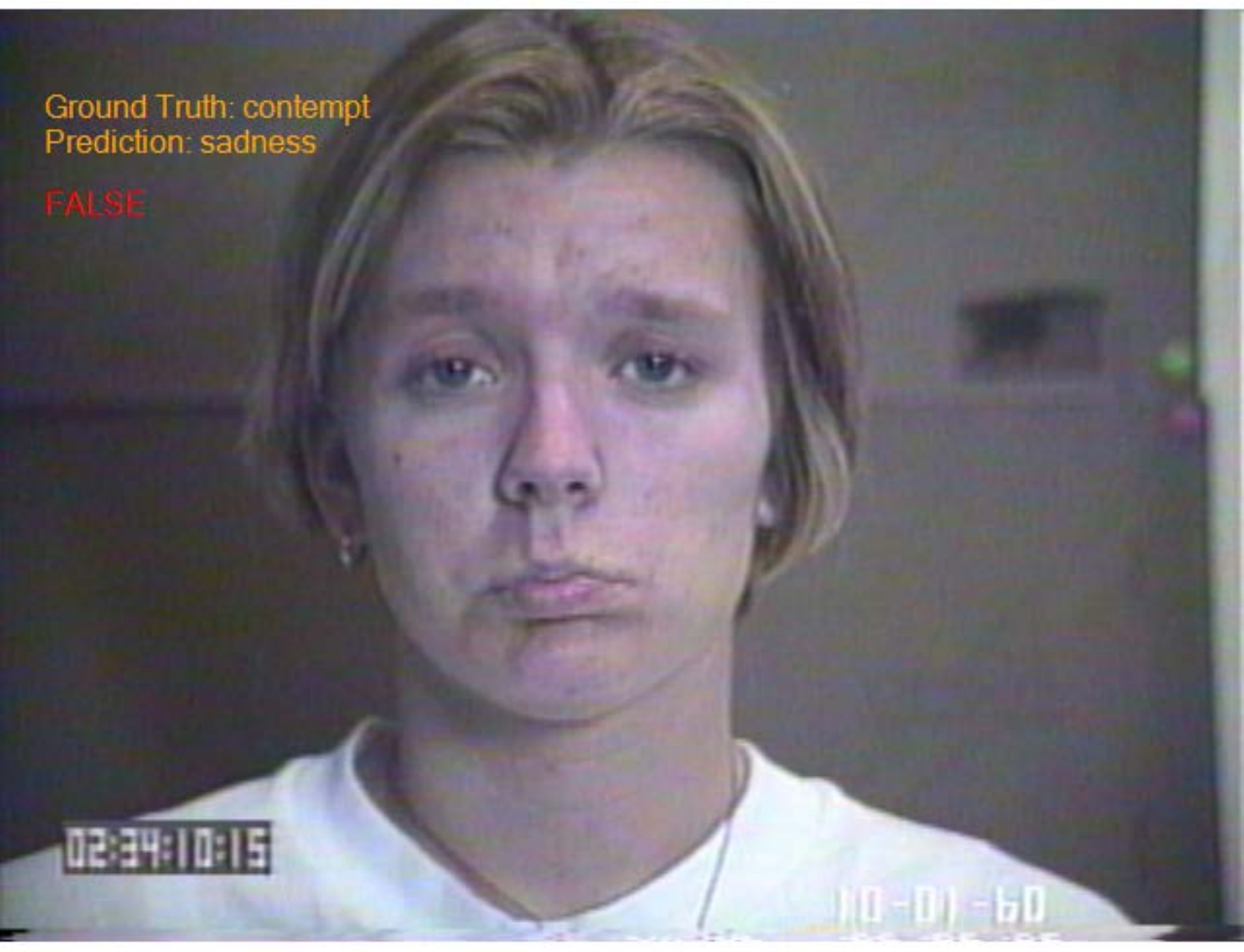}}}
	\subfloat[][Fear - 39\%\\ GT: Sadness - 31\%]{{\includegraphics[width=0.25\columnwidth]{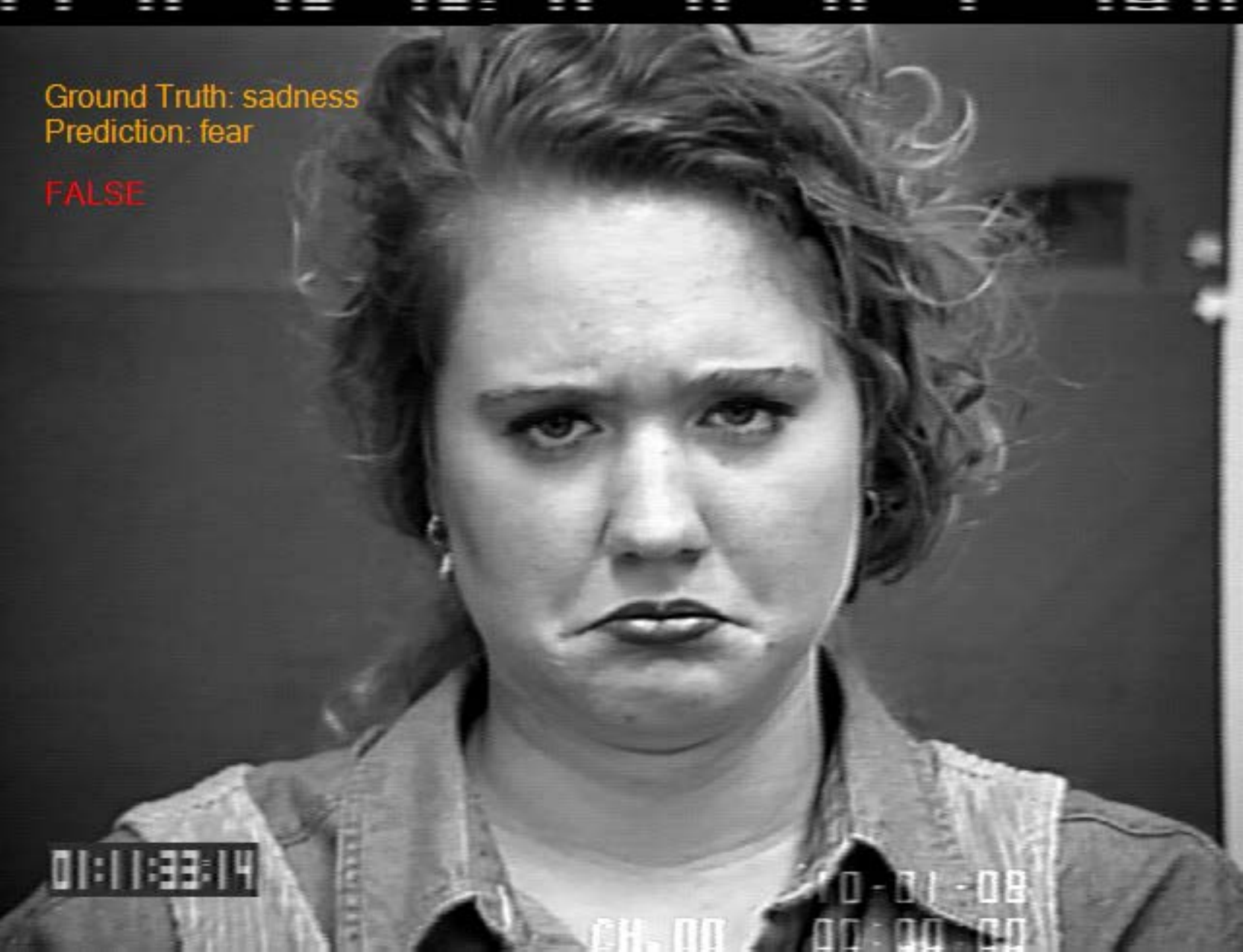}}}
	\subfloat[][Sadness - 81\% \\ GT: Anger - 19\%]{{\includegraphics[width=0.25\columnwidth]{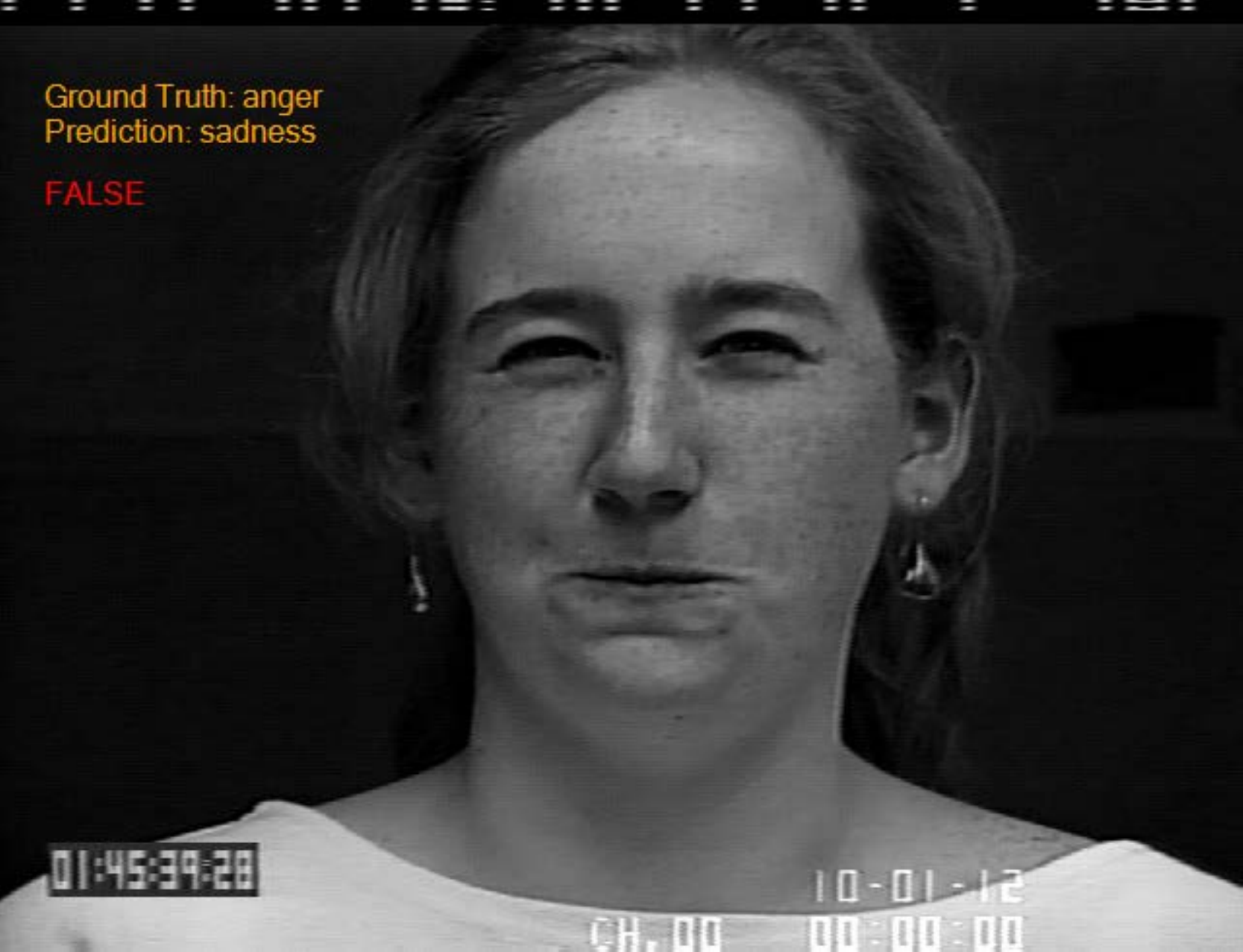}}}
	\subfloat[][Anger - 56\% \\ GT: Sadness - 43\%]{{\includegraphics[width=0.25\columnwidth]{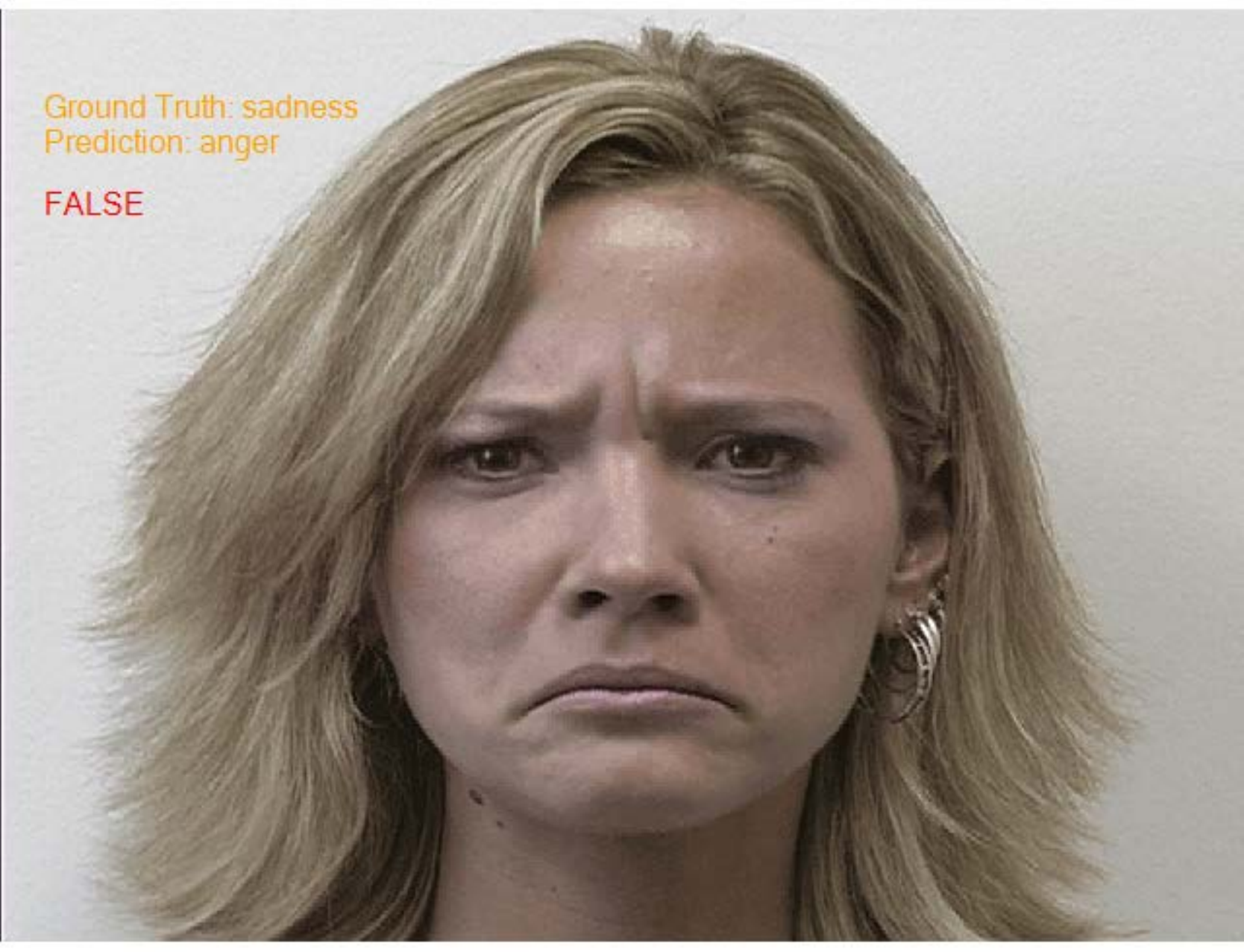}}}
	
	\vspace{-2mm}
	\caption{
		Posterior probabilities for classes are provided in percentages.
		For the incorrectly classified images, ground truth and its posterior probability is indicated in the second line.
	}
	\label{fig:qual}
\end{figure}

We tested the selected features for classification using 10-fold cross-validation.
The feature vector is $L2$ normalized and a one-against-one multiclass SVM with RBF kernel is used as the classifier.
See Table~\ref{tab:confmat} for the confusion matrix.
Accuracy is 88.7\%, and the mean of accuracies for individual classes is 82.4\%.
See Figure~\ref{fig:qual} for examples of classifications and respective posterior probabilities.

\begin{table}
	\caption{Classification accuracies using different geometric features in CK+.}
	\centering
	\resizebox{0.72\columnwidth}{!}
	{
		\tabulinesep=1mm
		\begin{tabu}{cccc}
			\tabucline[1.5pt]{-}
			& SPTS~\cite{Lucey:2010} & Geometry Features~\cite{Chen:2015} & Proposed \\
			\tabucline[1.5pt]{-}
			Anger & 0.35 & \textbf{0.89} & 0.78 \\
			Contempt & 0.25 & 0.39 & \textbf{0.64} \\
			Disgust & 0.68 & 0.90 & \textbf{0.93} \\
			Fear & 0.22 & 0.36 & \textbf{0.80} \\
			Happiness & 0.98 & \textbf{0.99} & \textbf{0.99} \\
			Sadness & 0.04 & \textbf{0.64} & \textbf{0.64} \\
			Surprise & \textbf{1} & 0.99 & \textbf{1} \\
			\tabucline[1.5pt]{-}
			Total & 0.665 & 0.847 & \textbf{0.887} \\
			\tabucline[1.5pt]{-}
		\end{tabu}
	}
	\label{tab:results}
\end{table}

SPTS features are the horizontal and vertical displacements of individual facial landmarks after any similarity transformation is rectified~\cite{Lucey:2010}.
To compute Geometry Features, facial landmarks are organized in a triangle mesh~\cite{Chen:2015}.
The edge lengths and angles of the triangles in the mesh change with facial expressions.
These changes are used as a type of geometric feature.
See Table~\ref{tab:results} for a comparison of classification accuracies using different geometric features.
The proposed geometric feature outperforms SPTS and Geometry Features for nearly all emotions.

Note that we have deliberately limited our comparison to geometric features.
It is common practice in the literature to combine geometric features with appearance-based features to boost performance.
The combination of SPTS and CAPP yield 88.4\% classification accuracy~\cite{Lucey:2010}, and the combination of Geometry Features and HOG from three orthogonal planes yield 93.6\% classification accuracy~\cite{Chen:2015}.

\section{Conclusion}

Geometric and appearance-based features tend to capture different representations of facial expressions, hence work well together.
Consequently, improvements for either of these feature types will be beneficial for facial expression recognition systems.
In this study, we proposed geometric features derived from landmark pairs, including many non-descriptive and redundant ones.
Instead of using this feature vector directly or applying a dimension reduction method, we used sequential forward selection to find a descriptive subset.

The selected spatial features yield 88.7\% recognition accuracy and surpass other purely geometric features in the literature.
To obtain better results, the selection can be done in an extended feature set, including many additional geometric and appearance-based features.
A feature selection algorithm that searches for a larger part of the feature subset space is also expected to improve our results.

\bibliographystyle{IEEEbib}
\bibliography{refs}

\end{document}